\documentclass[sigconf,nonacm]{acmart}
\usepackage{amsmath}
\usepackage{adjustbox}
\usepackage{multirow}
\usepackage{colortbl}
\usepackage{subfigure}

\usepackage{xspace}

\newcommand{\bfA}{\mathbf{A}}

\newcommand{\scaled}{S\MakeLowercase{ca}L\MakeLowercase{ed}\xspace}
\newtheorem{defn}{Definition}

\newcommand{\darkgray}{\cellcolor[rgb]{0.7,0.7,0.7}}
\newcommand{\lightgray}{\cellcolor[rgb]{0.86,0.86,0.86}}
\AtBeginDocument{%
  }

\begin{document}


\title{Sampling Enclosing Subgraphs for Link Prediction}


\author{Paul Louis}
\email{paul.louis@ontariotechu.net}
\affiliation{%
  \institution{Ontario Tech University}
  \city{Oshawa}
  \state{Ontario}
  \country{Canada}
}

\author{Shweta Ann Jacob}
\email{shweta.jacob@ontariotechu.net}
\affiliation{%
  \institution{Ontario Tech University}
  \city{Oshawa}
  \state{Ontario}
  \country{Canada}
}

\author{Amirali Salehi-Abari}
\email{abari@ontariotechu.ca}
\affiliation{%
  \institution{Ontario Tech University}
  \city{Oshawa}
  \state{Ontario}
  \country{Canada}
}


\begin{abstract}
 Link prediction is a fundamental problem for graph-structured data (e.g., social networks, drug side-effect networks, etc.). Graph neural networks have offered robust solutions for this problem, specifically by learning the representation of the subgraph enclosing the target link (i.e., pair of nodes). However, these solutions do not scale well to large graphs as extraction and operation on enclosing subgraphs are computationally expensive, especially for large graphs. This paper presents a scalable link prediction solution, that we call \scaled,  which utilizes sparse enclosing subgraphs to make predictions. To extract sparse enclosing subgraphs, \scaled takes multiple random walks from a target pair of nodes, then operates on the sampled enclosing subgraph induced by all visited nodes. By leveraging the smaller sampled enclosing subgraph, \scaled can scale to larger graphs with much less overhead while maintaining high accuracy. \scaled further provides the flexibility to control the trade-off between computation overhead and accuracy. Through comprehensive experiments, we have shown that \scaled can produce comparable accuracy to those reported by the existing subgraph representation learning frameworks while being less computationally demanding.
\end{abstract}

\settopmatter{printfolios=true}
\maketitle

\section{Introduction}

Graph-structured data such as user interactions, collaborations, protein-protein interactions, drug-drug interactions are prevalent in natural and social sciences. \textit{Link prediction}---a fundamental problem on graph-structured data---intends to quantify the likelihood of a link (or interaction) occurring between a pair of nodes (e.g., proteins, drugs, etc.). Link prediction has many diverse applications such as predicting drug side effects, drug-repurposing \cite{gysi2021network}, understanding molecule interactions \cite{huang2020skipgnn}, friendship recommendation \cite{chen2020friend}, and recommender systems \cite{ying2018graph}. 

Many solutions to link prediction problem \cite{liben2007link, lu2011link, martinez2016survey, wang2015link, kumar2020link} has been proposed ranging from simple heuristics (e.g., common neighbors, Adamic-Adar \cite{adamic2003friends}, Katz \cite{katz1953new}) to \emph{graph neural networks (GNNs)} \cite{kipf2016variational, zhang2018link, pan2022neural, hao2020inductive, cai2020multi, cai2021line}. Among these solutions, GNNs \cite{hamilton2020graph,wu2020comprehensive,zhou2020graph} have emerged as the widely-accepted and successful solution for learning rich latent representations of graph data to tackle link prediction problems. The early GNNs focused on \emph{shallow encoders} \cite{perozzi2014deepwalk, grover2016node2vec} in which the latent nodes' representations was first learnt through a sequence of random walks, and then a likelihood of a link is determined by combining its two-end nodes' latent representations. However, these shallow encoders were limited by not incorporating nodal features and their incompatibility with \emph{inductive settings} as they require that all nodes are present for training. These two challenges were (partially) addressed with the emergence of \emph{message-passing graph neural networks} \cite{kipf2017semi, hamilton2017inductive, xu2018how}. These advancements motivate the research on determining and extending the expressive power of GNNs \cite{you2021identity, bevilacqua2022equivariant, you2019position, zhang2018end, zeng2021decoupling, fey2021gnnautoscale} for all downstream tasks of link prediction, node classification, and graph classification. For link prediction, subgraph-based representation learning (SGRL) methods \cite{zhang2018link,li2020distance,pan2022neural, cai2020multi, cai2021line}---by learning the enclosing subgraphs around the two-end nodes rather than independently learning two end-node's embedding---have improved GNNs expressive power, and offered state-of-the-art solutions. However, these solutions suffer from the lack of scalability, thus preventing them to be applied to large-scale graphs. This is primarily due to the computation overhead in extracting, preprocessing, and learning (large) enclosing subgraphs for any pair of nodes. We focus on addressing this scalability issue.
\vskip 1.5mm
\noindent \textbf{Contribution.} We introduce \emph{\textbf{S}ampling En\textbf{c}losing Subgr\textbf{a}ph  for \textbf{L}ink Pr\textbf{ed}iction (\scaled)} to extend SGRL methods and enhance their scalability. The crux of \scaled is to sample enclosing subgraphs using a sequence of random walks. This sampling reduces the computational overhead of large subgraphs while maintaining their key structural information. $\scaled$ can be integrated into any GNN, and also offers parallelizability and model compression that can be exploited for large-scale graphs. Furthermore, the two hyperparameters, walk length and number of walks, in \scaled provides a way to control the trade-off between scalability and accuracy, if needed. Our extensive experiments on real-world datasets demonstrate that \scaled produces comparable results to the state-of-the-art methods (e.g, SEAL \cite{zhang2018link}) in link prediction, but requiring magnitudes less training data, time, and memory. \scaled combines the benefits of SGRL framework and random walks for link prediction. 
\vskip 1.5mm
\noindent \textbf{Other related work.} Graph neural networks have benefited from sampling techniques (e.g, node sampling \cite{hamilton2017inductive,chen2018fastgcn, zou2019layer} and graph sampling \cite{zeng2019graphsaint}) to make training and inference more scalable on large graphs. Also, historical embeddings \cite{chen2018stochastic, fey2021gnnautoscale} have shown promises in speeding up the aggregation step of GNNs for the downstream task of node classification \cite{chen2018fastgcn, chiang2019cluster, fey2021gnnautoscale}. However, little attention is given to scaling up of SGRL methods for link prediction; except a few exceptions \cite{yin2022algorithm, zeng2021decoupling}, which do not benefit from GNNs for subgraph learning \cite{yin2022algorithm} or is restricted to learning on  individual nodes \cite{zeng2021decoupling}.
\section{Link Prediction} 
We consider an undirected graph $G = (V, E, \bfA)$ where $V=[n]$ is the set of $n$ nodes (e.g., individuals, proteins, etc), $E \subseteq V\times V$ represents the edge set (e.g., friendship relations or protein-to-protein interactions) and the tensor $\bfA \in \mathbb{R}^{n\times n \times d}$ contains all nodes' attributes (e.g., user profiles) and edges' attributes (e.g, the strength or type of interactions). For each node $v \in V$, its attributes (if any) are stored in the diagonal component $\bfA_{vv.}$ while the off-diagonal component $\bfA_{uv.}$ can have the attributes of an edge $(u,v)$ if  $(u,v) \in E$; otherwise $\bfA{_{uv.}} = \mathbf{0}$. 


\vskip 1.5mm
\noindent \textbf{Link Prediction Problem}. Our goal in link prediction is to infer the presence or absence of an edge between a pair of \emph{target nodes} given the observed tensor $\bfA$. The learning problem is to find a \emph{likelihood (or scoring) function} $f$ such that it assigns \emph{interaction likelihood (or score)} $\hat{A}_{uv}$ to each target pair of nodes $(u,v) \notin E$, whose relationships to each other are not observed. Larger $\hat A_{uv}$ indicates a higher chance of $(u,v)$ forming a link or missing a link. The function $f$ can be formulated as $\hat{A}_{uv}$ = $f(u,v,\bfA |\boldsymbol{\theta})$ with $\boldsymbol{\theta}$ denoting the model parameters. Most link prediction methods differ from each other in the formulation of the likelihood function $f$ and its assumptions. The function $f$ can be some parameter-free predefined heuristics \cite{adamic2003friends,page1999pagerank,katz1953new} or learned by a graph neural network \cite{kipf2017semi,hamilton2017inductive,velickovic2018graph, xu2018how} or any other deep learning framework  \cite{yin2022algorithm}. The likelihood function formulation also varies based on its computation requirement on the maximum hop of neighbors of target nodes. For example, \emph{first-order heuristics} (e.g., common neighbors and preferential attachment \cite{barabasi1999emergence}) only require the direct neighbors while graph neural networks methods \cite{kipf2016variational,hamilton2017inductive} and high-order heuristics (e.g., Katz \cite{katz1953new}, rooted PageRank \cite{brin2012reprint}) require knowledge of the entire graph.



\begin{figure*}[tb]
  \centering
  \subfigure[Original graph with target nodes $u, v$.\label{original_graph} ]{\includegraphics[scale=0.58]{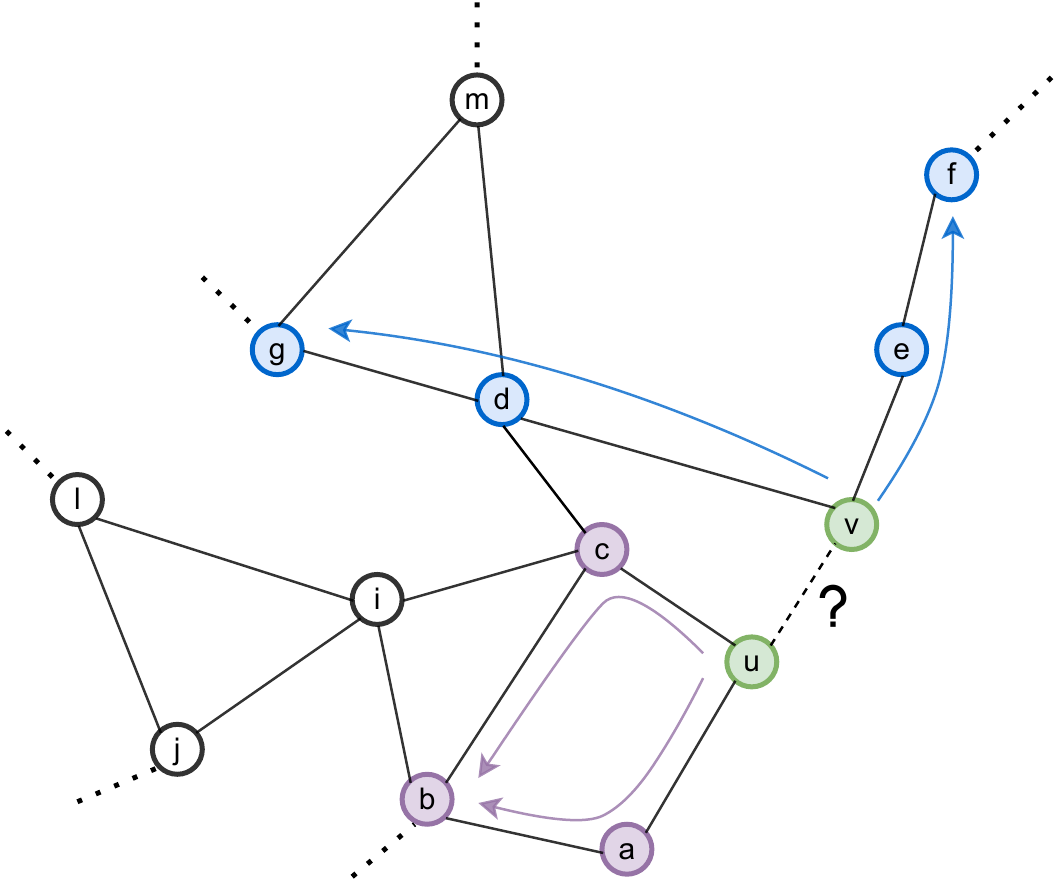}}\quad\quad\quad
  \subfigure[Induced subgraph with labels.\label{induced_subgraph}]{\includegraphics[scale=0.56]{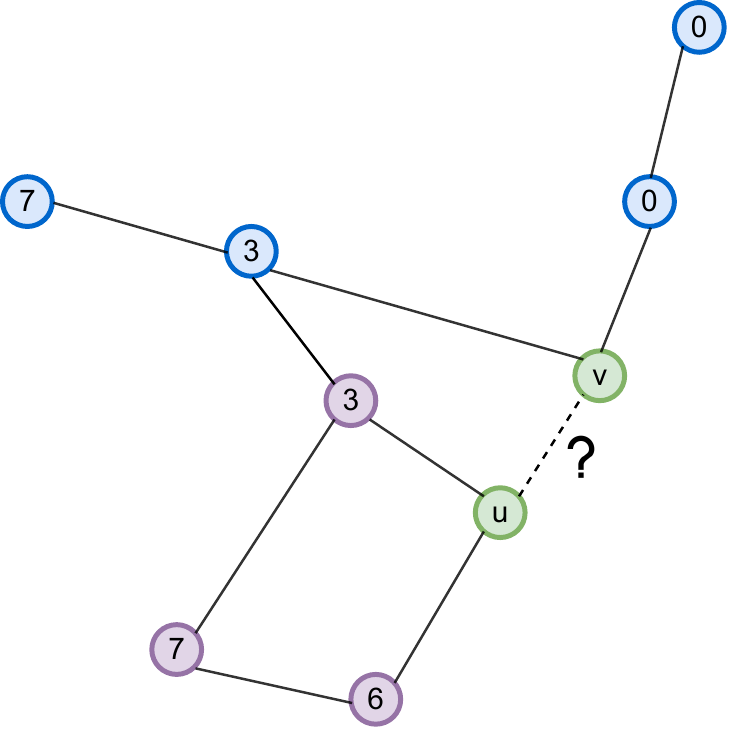}}
  \caption{The \scaled model: (a) the random walks from target nodes $u$ (blue arrows) and $v$ (purple arrows) ; and (b) the induced sampled enclosing subgraph with DRNL labels.}
  \label{fig:scaled}
\end{figure*}

\section{The S\lowercase{ca}L\lowercase{ed} model}
After describing the SEAL link prediction model and its variants, we detail how our proposed \emph{\scaled} model extends these models to maintain their prediction power but offer better scalability.




\vskip 1.5mm
\noindent \textbf{SEAL and its variants}. Rather than learning the target nodes' embeddings independently (as with Graph Convolutional Network \cite{kipf2017semi} or GraphSAGE \cite{hamilton2017inductive}), SEAL \cite{zhang2018link} focuses on learning the \emph{enclosing subgraph} of a pair of target nodes to capture their relative positions to each other in the graph: 

\begin{defn}[Enclosing Subgraph \cite{zhang2018link}]
Given a graph $G$, the $h$-hop enclosing subgraph around target nodes (u,v) is the subgraph $G_{uv}^{h}$ induced from $G$ with the set of nodes $\{j \| d(j,x) \leq h\ or\ d(j,y) \leq h \}$, where $d(i,j)$ is the geodesic distance between node $i$ and $j$.
\label{def:se}
\end{defn}

In SEAL, for each pair of the target nodes $(u,v)$, their enclosing subgraph $G_{uv}^{h}$ is found and extracted with two $h$-hop Breadth-First Search (BFS), where each BFS starts from $u$ and $v$. The nodes in the extracted enclosing subgraph are also augmented with labels indicating their distances to the target pair of nodes using the \textit{Double-Radius Node Labeling} (DRNL) hash function \cite{zhang2018link}:
\begin{equation}
    DRNL(x,G_{uv}^{h}) = 1 + min\left(d_{xu}, d_{xv}\right) + \lfloor d'/2 \rfloor \lceil d'/2 - 1 \rceil
    \label{drnl_eq}, 
\end{equation}
where $x$ represents the nodes in the subgraph $G_{uv}^{h}$, $d_{xu}$ is the geodesic distance of $x$ to $u$ in $G_{uv}^{h}$ when node $v$ is removed, and $d' = d_{xu} + d_{xv}$. Note that  the distance of $x$ to each target node $u$ is calculated in isolation by removing the other target node $v$ from the subgraph. The target nodes are labeled as 1 and a node with $\infty$ distance to at least one of the target nodes is given the label $0$. Each node label is then represented by its one-hot encoding, and expands the initial node features, if any. The subgraph $G_{uv}^{h}$ along with the augmented nodal features is fed into a graph neural network, which predicts the presence or absence of the edge. In SEAL, the link prediction is treated as a binary classification over the enclosing subgraphs by determining if the enclosing subgraph will be closed by a link between the pair of target nodes or not.  Thus, SEAL uses a graph pooling mechanism (e.g., SortPooling \cite{zhang2018end}) to compute the enclosing subgraph representation for the classification task. 
Other variants of SEAL (e.g., DE-GNN \cite{li2020distance} and WalkPool \cite{pan2022neural}) have replaced either its DRNL labeling method \cite{li2020distance} or graph aggregation method \cite{li2020distance} with some other alternatives to improve its expressivness power. However, SEAL and these variants all suffer from the scalability issue as the subgraph size grows exponentially with the hop size $h$, and large-degree nodes (e.g., celebrities) possess very large enclosing subgraphs even for a small $h$.  To address these scalablity issues, we propose \emph{\textbf{S}ampling En\textbf{c}losing Subgr\textbf{a}ph  for \textbf{L}ink Pr\textbf{ed}iction (\scaled)}. 

\vskip 1.5mm
\noindent \textbf{\scaled}. Observing that the computational bottleneck of SEAL and its variants originates from the exponential growth and the size of enclosing subgraphs, we propose \emph{Sampled Enclosing Subgraphs} with more tractable sizes:

\begin{defn}[Random Walk Sampled Enclosing Subgraph]
Given a graph $G$, the random-walk sampled $h$-hop enclosing subgraph around target nodes (u,v) is the subgraph $\hat{G}_{uv}^{h,k}$ induced from $G$ with the set of nodes $\hat{V}^{h,k}_{uv} \in W_u^{h,k} \cup W_v^{h,k}$, where $W_i^{h,k}$ is the set of nodes visited by $k$ many h-length random-walk(s) from node $i$. 
\label{def:ses}
\end{defn}

Figure \ref{induced_subgraph} illustrates sampled enclosing subgraph of the target pair of $(u, v)$ for the original graph in Figure \ref{original_graph}, where $h=2$ and $k=2$. Here, $W_v^{h,k} = \{v, d, e, f, g\}$ and $W_u^{h,k} = \{u, a, b, c\}$, resulting in  $\hat{V}^{h,k}_{uv} = \{a, b, c, d, e, f, g, u, v\}$. The included subgraph in Figure \ref{induced_subgraph} contains all nodes and edges between nodes in $\hat{V}^{h,k}_{uv}$.

Comparing Definitions \ref{def:se} and \ref{def:ses}, a few important observations can be made: (i) the sampled enclosing subgraph  $\hat{G}_{uv}^{h,k}$ is the subgraph of the enclosing subgraph $G_{uv}^h$, as the $h$-length random walks can not reach a node further than $h$-hop away from the starting node; (ii) the size of the sampled subgraph is bounded to $O(hk)$ and controlled by these two parameters compared to the exponential growth of enclosing subgraphs with $h$ in Definition \ref{def:se}. 
\scaled, by replacing the dense enclosing subgraphs with their sparse (sub)subgraphs, offers scalability, while still providing flexibility to control the extent of sparsity and scalability with its sampling parameters $h$ and $k$. 

The \scaled model can use any labeling trick (e.g., DRNL, zero-one labeling, etc.) \cite{zhang2021labeling} to encode the distances between target nodes and other nodes in the sampled subgraphs; see Figure \ref{induced_subgraph} for an example. Similar to SEAL, the one-hot encoding of the distance labels along with the nodal features (if any) of the nodes in the sampled subgraph are fed into a graph neural network with graph pooling operation (e.g., DGCNN with SortPooling operation \cite{zhang2018end}) for the classification task. The \scaled model offers easy plug-and-play modularity into most graph neural networks (e.g., GCN \cite{kipf2017semi}, GIN \cite{xu2018how}, GraphSAGE \cite{hamilton2017inductive}, DGCNN \cite{zhang2018end}, etc.), and can also be used alongside any regularization technique or loss function.

Although random walks have been used in unsupervised latent learning of graph data \cite{perozzi2014deepwalk, grover2016node2vec}, \scaled has used them differently for sparsifying the enclosing subgraphs to enhance scalability. This random-walk subgraph sampling technique can be incorporated into any other subgraph representation learning task to improve scalability by increasing sparsity of the subgraph. This technique does not incur much computational overhead, can be viewed as a preprocessing step, and can benefit from parallelizability. Furthermore, random-walk sampled subgraphs have controllable size and and do not grow exponentially with $h$.  

\begin{table}
\centering
\begin{tabular}{lllll} 
\toprule
\textbf{Dataset} & \textbf{\# Nodes} & \textbf{\# Edges} & \textbf{Avg. Deg.} & \textbf{\# Features}  \\ 
\hline
USAir            & 332               & 2126              & 12.81              & NA                    \\
Celegans         & 297               & 2148              & 14.46              & NA                    \\
NS               & 1461              & 2742              & 3.75               & NA                    \\
Router           & 5022              & 6258              & 2.49               & NA                    \\
Power            & 4941              & 6594              & 2.67               & NA                    \\
Yeast            & 2375              & 11693             & 9.85               & NA                    \\
Ecoli            & 1805              & 14660             & 16.24              & NA                    \\
PB               & 1222              & 16714             & 27.36              & NA                    \\ 
\hline
Cora             & 2708              & 5429              & 4                  & 1433                  \\
CiteSeer         & 3327              & 4732              & 2.84               & 3703                  \\
\bottomrule
\end{tabular}
\caption{The statistics of experimented datasets.} 
\label{table:dataset_comparison}
\end{table}
\begin{table*}[tb]
\centering
\setlength{\extrarowheight}{0pt}
\addtolength{\extrarowheight}{\aboverulesep}
\addtolength{\extrarowheight}{\belowrulesep}
\setlength{\aboverulesep}{0pt}
\setlength{\belowrulesep}{0pt}
\scalebox{0.87}{
\begin{tabular}{lllllllllll} 
\toprule
\textbf{Model}  & \multicolumn{1}{c}{\textbf{USAir}}               & \multicolumn{1}{c}{\textbf{Celegans}}            & \multicolumn{1}{c}{\textbf{NS}}                  & \multicolumn{1}{c}{\textbf{Router}}              & \multicolumn{1}{c}{\textbf{Power}}               & \multicolumn{1}{c}{\textbf{Yeast}}               & \multicolumn{1}{c}{\textbf{Ecoli}}               & \multicolumn{1}{c}{\textbf{PB}}                  & \multicolumn{1}{c}{\textbf{CiteSeer}}            & \multicolumn{1}{c}{\textbf{Cora}}              \\ 
\hline
CN              & 93.02 $\pm$ 1.16                                     & 83.46 $\pm$ 1.22                                     & 91.81 $\pm$ 0.78                                     & 55.48 $\pm$ 0.61                                     & 58.10 $\pm$ 0.53                                     & 88.75 $\pm$ 0.70                                     & 92.76 $\pm$ 0.70                                     & 91.35 $\pm$ 0.47                                     & 65.90 $\pm$ 0.99                                     & 71.47 $\pm$ 0.70                                                           \\
AA              & 94.34 $\pm$ 1.31                                     & 85.26 $\pm$ 1.14                                     & 91.83 $\pm$ 0.75                                     & 55.49 $\pm$ 0.61                                     & 58.10 $\pm$ 0.54                                     & 88.81 $\pm$ 0.68                                     & 94.61 $\pm$ 0.52                                     & 91.68 $\pm$ 0.45                                     & 65.91 $\pm$ 0.98                                     & 71.54 $\pm$ 0.72                                                                 \\
PPR             & 88.61 $\pm$ 2.01                                     & 85.24 $\pm$ 0.64                                     & 91.95 $\pm$ 1.11                                     & 39.88 $\pm$ 0.51                                     & 63.09 $\pm$ 1.90                                     & 91.65 $\pm$ 0.74                                     & 89.77 $\pm$ 0.48                                     & 86.93 $\pm$ 0.54                                     & 73.85 $\pm$ 1.39                                     & 82.58 $\pm$ 1.13                                                                          \\
GCN             & 88.03 $\pm$ 2.84                                     & 81.58 $\pm$ 1.42                                     & 91.48 $\pm$ 1.28                                     & 83.99 $\pm$ 0.64                                     & 67.51 $\pm$ 1.21                                     & 90.80 $\pm$ 0.95                                     & 90.82 $\pm$ 0.56                                     & 90.92 $\pm$ 0.72                                     & 86.66 $\pm$ 1.02                                     & 89.36 $\pm$ 0.99                                                             \\
SAGE            & 85.64 $\pm$ 1.60                                     & 74.68 $\pm$ 4.46                                     & 91.02 $\pm$ 2.58                                     & 67.33 $\pm$ 10.49                                    & 65.77 $\pm$ 1.06                                     & 88.08 $\pm$ 1.63                                     & 87.12 $\pm$ 1.14                                     & 86.75 $\pm$ 1.83                                     & 84.13 $\pm$ 1.07                                     & 85.86 $\pm$ 1.27                                                               \\
GIN             & 88.93 $\pm$ 2.04                                     & 73.60 $\pm$ 3.17                                     & 82.16 $\pm$ 2.70                                     & 75.74 $\pm$ 3.31                                     & 57.93 $\pm$ 1.28                                     & 83.51 $\pm$ 0.67                                     & 89.34 $\pm$ 1.45                                     & 90.35 $\pm$ 0.78                                     & 71.73 $\pm$ 4.11                                     & 71.77 $\pm$ 2.74                                               \\
MF              & 89.99 $\pm$ 1.74                                     & 75.81 $\pm$ 2.73                                     & 77.66 $\pm$ 3.02                                     & 69.92 $\pm$ 3.26                                     & 51.30 $\pm$ 2.25                                     & 86.88 $\pm$ 1.37                                     & 91.07 $\pm$ 0.39                                     & 91.74 $\pm$ 0.22                                     & 61.24 $\pm$ 3.96                                     & 60.68 $\pm$ 1.30                                                                 \\
n2v             & 86.27 $\pm$ 1.39                                     & 74.86 $\pm$ 1.38                                     & 90.69 $\pm$ 1.20                                     & 63.30 $\pm$ 0.53                                     & 72.58 $\pm$ 0.71                                     & 90.91 $\pm$ 0.58                                     & 91.02 $\pm$ 0.17                                     & 84.84 $\pm$ 0.73                                     & 74.86 $\pm$ 1.11                                     & 78.79 $\pm$ 0.75                                                               \\
SEAL            & \darkgray 97.39 $\pm$ 0.72       & \darkgray 90.71 $\pm$ 1.39       & \lightgray 98.65 $\pm$ 0.57 & \darkgray 95.70 $\pm$ 0.17       & \darkgray 84.73 $\pm$ 1.14       & \lightgray 97.48 $\pm$ 0.25 & \darkgray 97.88 $\pm$ 0.20       & \darkgray 95.08 $\pm$ 0.39       & \darkgray 88.50 $\pm$ 1.15       & \darkgray 90.66 $\pm$ 0.81                              \\
\textbf{ScaLed} & \lightgray 96.44 $\pm$ 0.93 & \lightgray 88.27 $\pm$ 1.17 & \darkgray 98.88 $\pm$ 0.50       & \lightgray 94.20 $\pm$ 0.50 & \lightgray 83.99 $\pm$ 0.84 & \darkgray 97.68 $\pm$ 0.17       & \lightgray 97.31 $\pm$ 0.14 & \lightgray 94.53 $\pm$ 0.57 & \lightgray 87.69 $\pm$ 1.67 & \lightgray 90.55 $\pm$ 1.18                                       \\
\bottomrule
\end{tabular}
}
\caption{Average AUCs with 5 random seeds. The best and second best are shaded in dark and light gray respectively.}
\vspace{-20pt}
\label{table:accuracy_results}
\end{table*}
\section{Experiments}

We run an extensive set of experiments to compare the prediction accuracy and computational efficiency of \scaled against a set of state-of-the-art methods for link prediction. We further analyze its hyperparameter sensitivity and its efficacy in improving subgraph sparsity.\footnote{Our code is implemented in PyTorch Geometric \cite{Fey/Lenssen/2019} and PyTorch \cite{paszke2019pytorch}. The link to GitHub repository is \url{https://github.com/venomouscyanide/ScaLed}. All our experiments are run on servers with 50 CPU cores, 377 GB RAM and 4$\times$NVIDIA GTX 1080 Ti 11GB GPUs.} 
\vskip 1mm
\noindent \textbf{Datasets.} We consider a set of homogeneous, undirected graph datasets  (see Table \ref{table:dataset_comparison}), which have been commonly subject to many other link prediction studies \cite{zhang2017weisfeiler, zhang2018link,cai2020multi, cai2021line, hao2020inductive, li2020distance, pan2022neural, wang2022equivariant} and are publicly available. Our datasets are categorized into \textit{non-attributed} and \textit{attributed} datasets where nodal features are absent or present in the dataset, respectively. The edges in each dataset are randomly split into 85\% training, 5\% validation, and 10\% testing datasets. Each dataset split is also augmented with random negative samples (i.e, absent links) at a 1:1 ratio for positive and negative samples.

\vskip 1.5mm \noindent \textbf{Baselines.} We compare our \scaled against a comprehensive set of baselines in four categories: 
heuristic, graph autoencoder (GAE), latent feature-based (LFB), and SGRL methods. For heuristic methods, we use common neighbors(CN), Adamic Adar(AA) \cite{adamic2003friends}, and Personalized PageRank (PPR). 
GAE baselines include GCN \cite{kipf2017semi}, GraphSAGE \cite{hamilton2017inductive}, and GIN \cite{xu2018how} encoders with a hadamard product of a pair of nodes' embedding as the decoder. The LFB methods consists of matrix factorization \cite{koren2009matrix} and node2vec \cite{grover2016node2vec} with the logistic classifier head. Our SGRL baseline is state-of-the-art method SEAL \cite{zhang2018link}. 

\vskip 1.5mm
\noindent \textbf{Setup.} 
GAE baselines have three hidden layers with dimensionality of 32. The nodal initial features, for non-attributed datasets, are set to one-hot indicators. In MF, the nodal latent feature has 32 dimensions for each node. MF uses a three-layered MLP with 32 hidden dimensions. For node2vec, we set sampling parameters $p=q=1$ and a dimensionality of $32$ for the node features. For SEAL, we set $h=2$ for non-attributed datasets and $h=3$ for attributed datasets. We also use a three-layered DGCNN with a hidden dimensionality of 32 for all datasets. 
For \scaled model, we set $k=20$ while $h$ and all other hyperparameters are set the same as that of SEAL for fair comparison. The learning rate is set to 0.0001 for SEAL and \scaled and 0.01 for node2vec, MF and GAE baselines. All learning models for both attributed and non-attributed datasets are trained for 50 epochs with a dropout of $0.5$ (except for node2vec without dropout) and Adam \cite{kingma2014adam} optimizer (except for node2vec with Sparse Adam). GAE baselines are trained by full-batch gradients; but others are trained with a batch size of 32. The detailed experimental setup used for each baseline and dataset is available in Github.\footnote{\url{https://github.com/venomouscyanide/ScaLed/tree/main/experiments}}

\vskip 1.5mm \noindent \textbf{Measurements.} For all models, we report the mean of area under the curve (AUC) of the testing data over 5 runs with 5 random seeds. For each model in each run, we test it against testing data with those parameters which achieve highest AUC of validation data over training with 50 epochs.  
For computational resource measurements, we also report average training-plus-inference time, allocated CUDA memory, model size, and number of parameters.

\begin{table}
\centering
\scalebox{0.75} {
\begin{tabular}{llrrrrrr} 
\toprule
\textbf{Dataset}             & \textbf{Model} & \textbf{Time} & \textbf{CUDA}   & \multicolumn{1}{c}{\textbf{Size}} & \textbf{Params.} & \textbf{\# Nodes} & \textbf{\# Edges}  \\ 
\hline
\multirow{2}{*}{USAir}       & SEAL           & 486           & 52 MB           & 2.04 MB                           & 0.533M           & 207               & 2910               \\
                             & ScaLed         & 446           & 11 MB           & 0.44 MB                           & 0.113M           & 40                & 518                \\ 
\hline
\multirow{2}{*}{Celegans}    & SEAL           & 473           & 47 MB           & 1.84 MB                           & 0.480M           & 206               & 2482               \\
                             & ScaLed         & 453           & 7 MB            & 0.45 MB                           & 0.117M           & 45                & 293                \\ 
\hline
\multirow{2}{*}{NS}          & SEAL           & 580           & 5 MB            & 0.22 MB                           & 0.056M           & 17                & 83                 \\
                             & ScaLed         & 572           & 3 MB            & 0.19 MB                           & 0.048M           & 12                & 55                 \\ 
\hline
\multirow{2}{*}{Router}      & SEAL           & 1330          & 38 MB           & 0.55 MB                           & 0.144M           & 82                & 253                \\
                             & ScaLed         & 1342          & 5 MB            & 0.27 MB                           & 0.683M           & 21                & 54                 \\ 
\hline
\multirow{2}{*}{Power}       & SEAL           & 1394          & 4 MB            & 0.22 MB                           & 0.056M           & 16                & 33                 \\
                             & ScaLed         & 1404          & 3 MB            & 0.20 MB                           & 0.052M           & 13                & 25                 \\ 
\hline
\multirow{2}{*}{Yeast}       & SEAL           & 2605          & 65 MB           & 1.38 MB                           & 0.362M           & 151               & 2438               \\
                             & ScaLed         & 2482          & 13 MB           & 0.39 MB                           & 0.101M           & 35                & 384                \\ 
\hline
\multirow{2}{*}{Ecoli}       & SEAL           & \textbf{6044} & 331 MB          & \textbf{10.22 MB}                 & \textbf{2.68M}   & \textbf{1166}     & 21075              \\
                             & ScaLed         & \textbf{3181} & 20 MB           & \textbf{0.50 MB}                  & \textbf{0.130M}  & \textbf{46}       & 790                \\ 
\hline
\multirow{2}{*}{PB}          & SEAL           & 6167          & \textbf{312 MB} & 6.41 MB                           & 1.68M            & 729               & \textbf{20981}     \\
                             & ScaLed         & 3649          & \textbf{15 MB}  & 0.57 MB                           & 0.149M           & 57                & \textbf{652}       \\ 
\hline
\multirow{2}{*}{CiteSeer}    & SEAL           & 1491          & 221 MB          & 0.97 MB                           & 0.253M           & 82                & 326                \\
                             & ScaLed         & 1044          & 49 MB           & 0.72 MB                           & 0.187M           & 22                & 63                 \\ 
\hline
\multirow{2}{*}{Cora}        & SEAL           & 1731          & 195 MB          & 1.78 MB                           & 0.466M           & 202               & 692                \\
                             & ScaLed         & 1199          & 27 MB           & 0.53 MB                           & 0.140M           & 32                & 88                 \\ 
                             
\midrule
\midrule

\multicolumn{2}{c}{\textbf{Maximum Ratio}}    & 1.90          & 20.80           & 20.44                           & 20.61            & 25.34             & 32.18              \\
\bottomrule
\end{tabular}
}
\caption{Average computational consumption of SEAL vs. \scaled  over five runs: runtime in seconds, max allocated CUDA and model size in Megabytes (MB), number of parameters in Millions.  The number of nodes and edges corresponds to the enclosing subgraphs. Maximum ratio corresponds to the maximum of ratio of SEAL's resource over \scaled's resource (in bold).}
\label{table:profile_results}
\end{table}
\vskip 1.5mm \noindent \textbf{Results: AUC.} 
Table \ref{table:accuracy_results} reports the average AUC over five runs for all datasets and link prediction models. In all attributed and non-attributed datasets, \scaled is ranked first or second among all baselines.  Also, \scaled gives very comparable results to SEAL or even outperforms SEAL in some datasets (e.g., NS and Yeast). This performance has also been achieved by order of magnitudes less resource consumption. 

\vskip 1.5mm \noindent \textbf{Results: Resource Consumption.} Table \ref{table:profile_results} reports the average consumption of resources over 5 runs for \scaled and SEAL. For all datasets, the average runtime of \scaled is much lower for larger datasets (e.g., Ecoli and PB), but slightly lower for small datasets (USAir and Celegans). For Ecoli and PB, \scaled gains speed up of 1.90$\times$ and 1.69$\times$ over SEAL, while using upto 20$\times$ less allocated GPU memory, model size and parameters. The sampled subgraphs in \scaled are sparser than that of SEAL (compare the number of nodes and edges in Table \ref{table:profile_results}). \scaled requires 7.86$\times$ and 5.17$\times$ less edges for Cora and CiteSeer, respectively. This compression can be upto 32.18$\times$ (see PB). The results in Tables \ref{table:accuracy_results} and Tables \ref{table:profile_results} confirm our hypothesis that \scaled is able to match the performance of SEAL with much less computational overhead. We even witness that \scaled has outperformed SEAL for NS and Yeast while consuming 1.51$\times$ and $6.35\times$ less edges in the sampled enclosing subgraphs. These results suggest that random walk based subgraph sampling is beneficial for the learning without compromising the accuracy. Finally, the results on larger and denser datasets such as Ecoli and PB indicates that the largest computational efficiency gains are achieved by \scaled on larger and denser datasets.
\begin{figure*}[htb]
  \centering
    \subfigure[AUC for USAir.\label{subfig1}]{\includegraphics[scale=0.29]{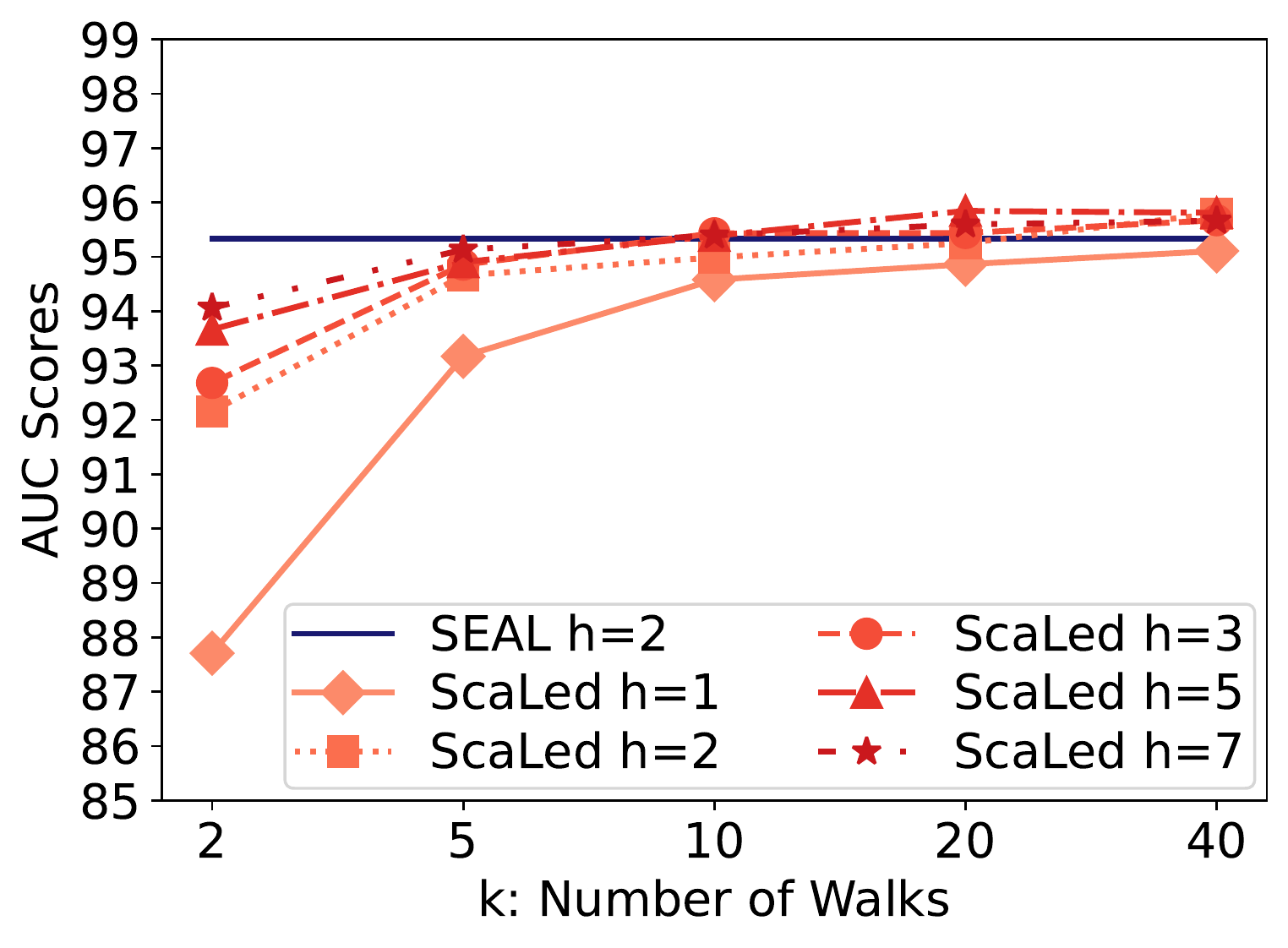}}\hspace{-4pt}
    \subfigure[Runtime for USAir.\label{runtimesubfig1}]{\includegraphics[scale=0.29]{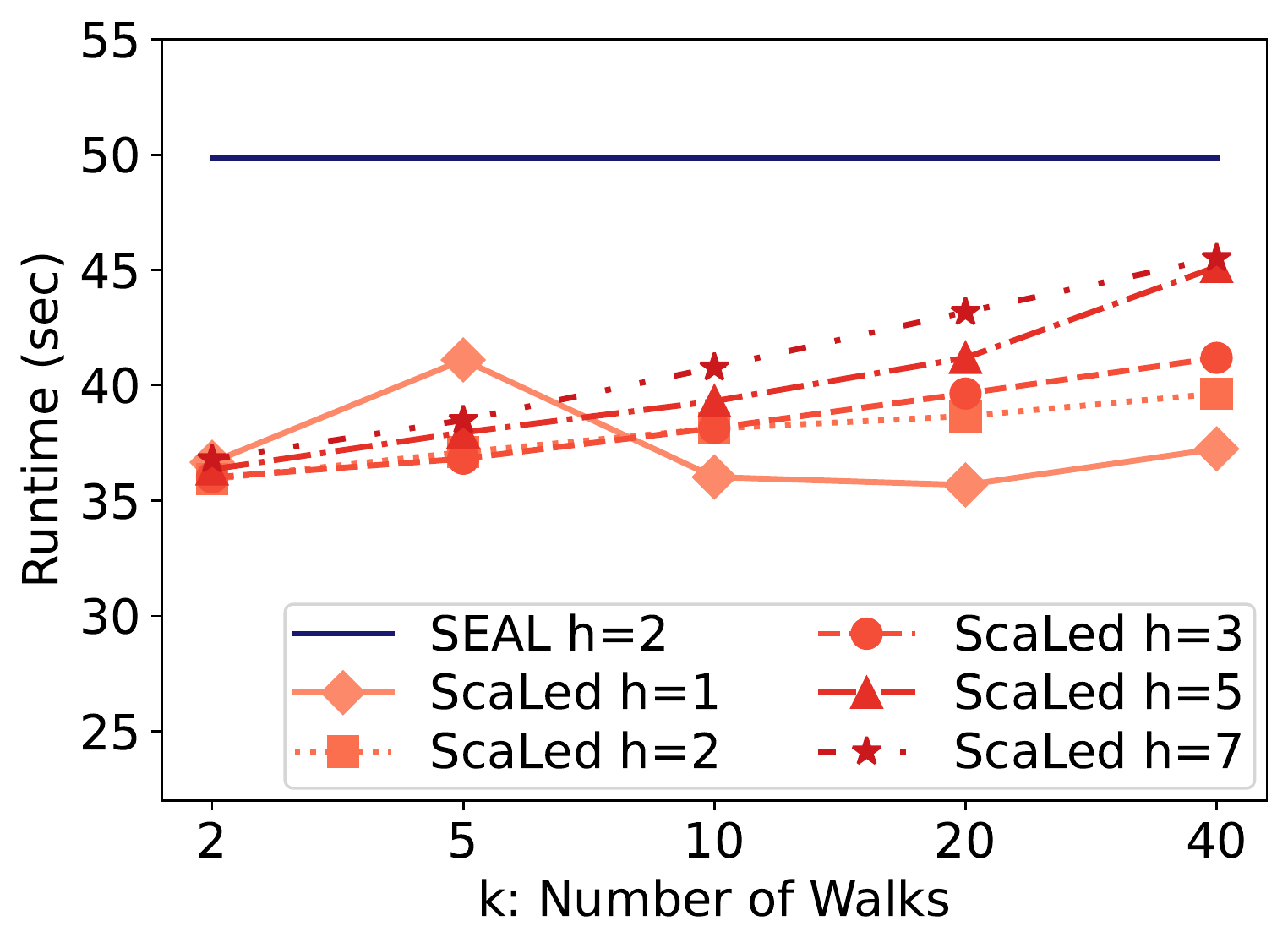}}\hspace{-4pt}
  \subfigure[AUC for Celegans.\label{subfig2}]{\includegraphics[scale=0.29]{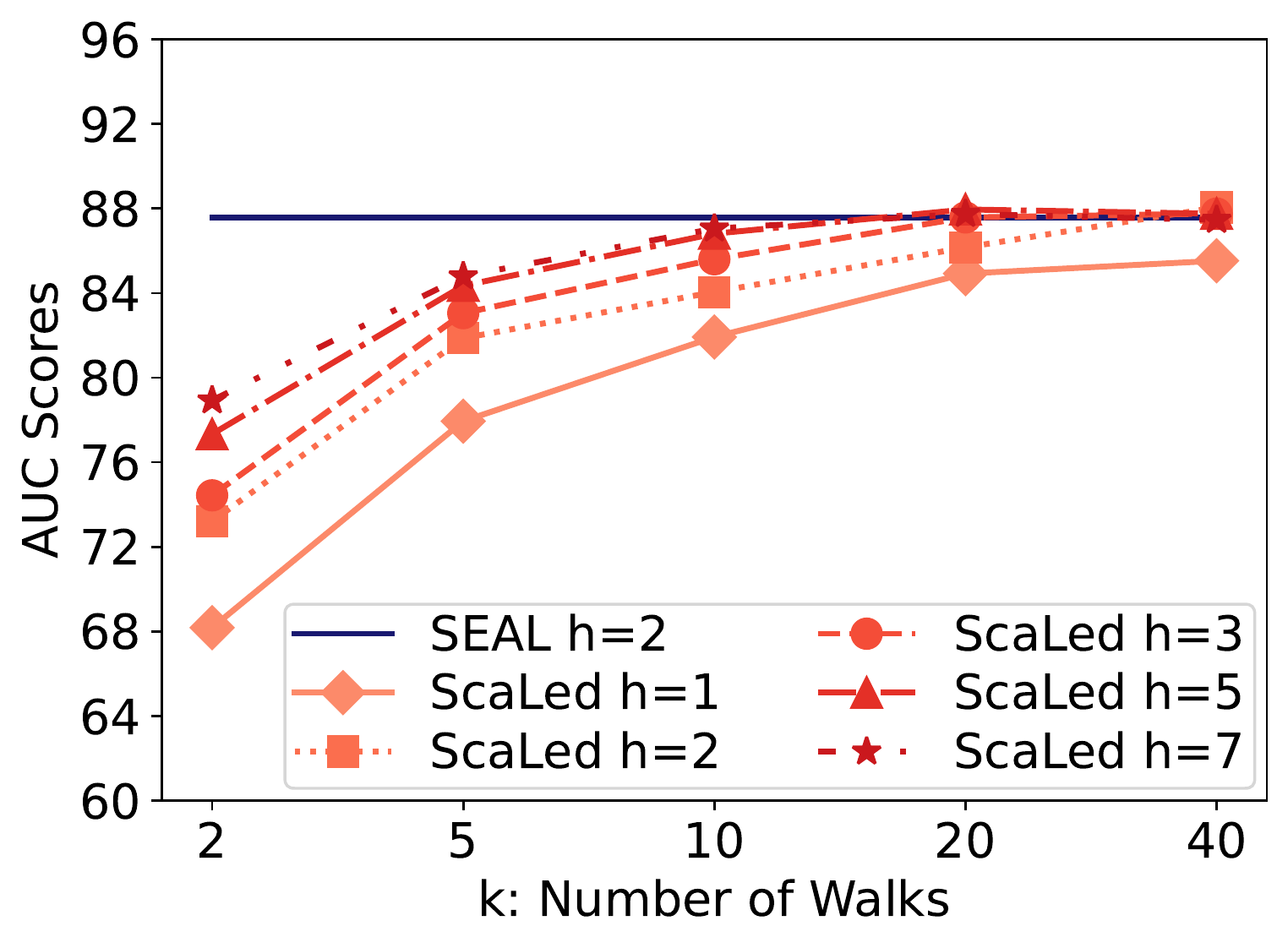}}\hspace{-4pt}
  \subfigure[Runtime for Celegans.\label{runtimesubfig2}]{\includegraphics[scale=0.29]{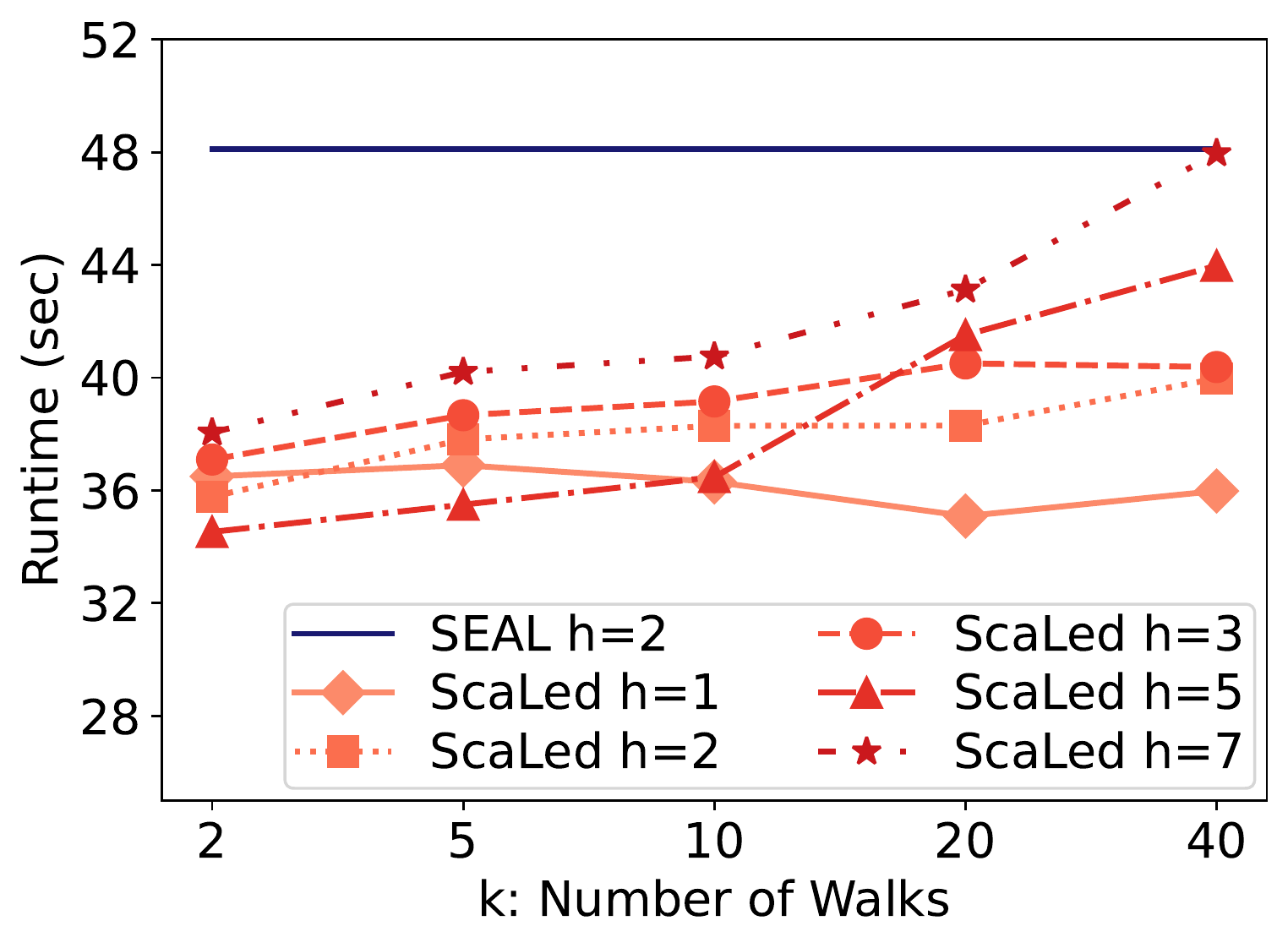}}\hspace{-4pt}
  \subfigure[AUC for NS.\label{subfig3}]{\includegraphics[scale=0.29]{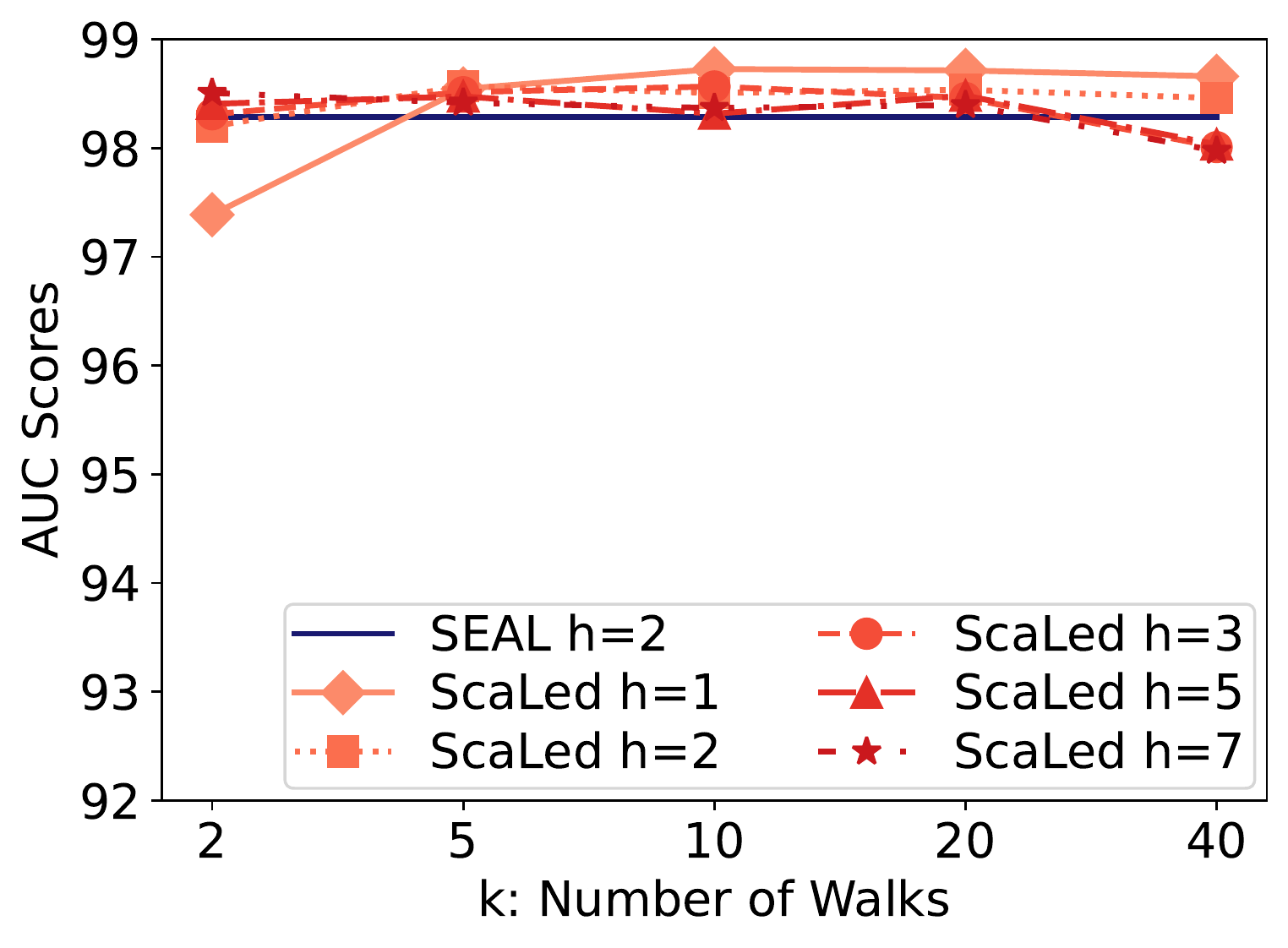}}\hspace{-4pt}
  \subfigure[Runtime for NS.\label{runtimesubfig3}]{\includegraphics[scale=0.29]{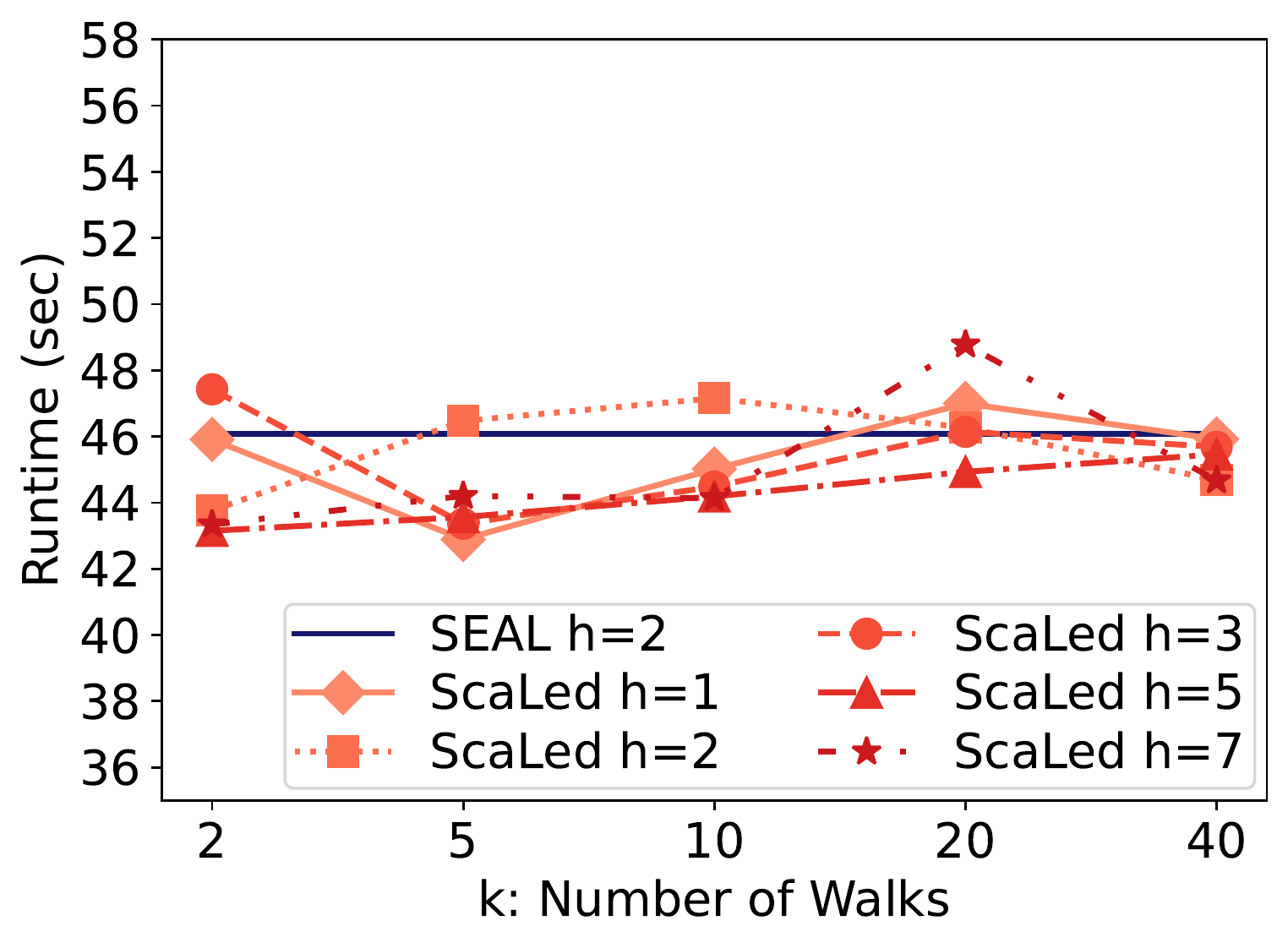}}\hspace{-4pt}
  \subfigure[AUC for Router.\label{subfig4}]{\includegraphics[scale=0.29]{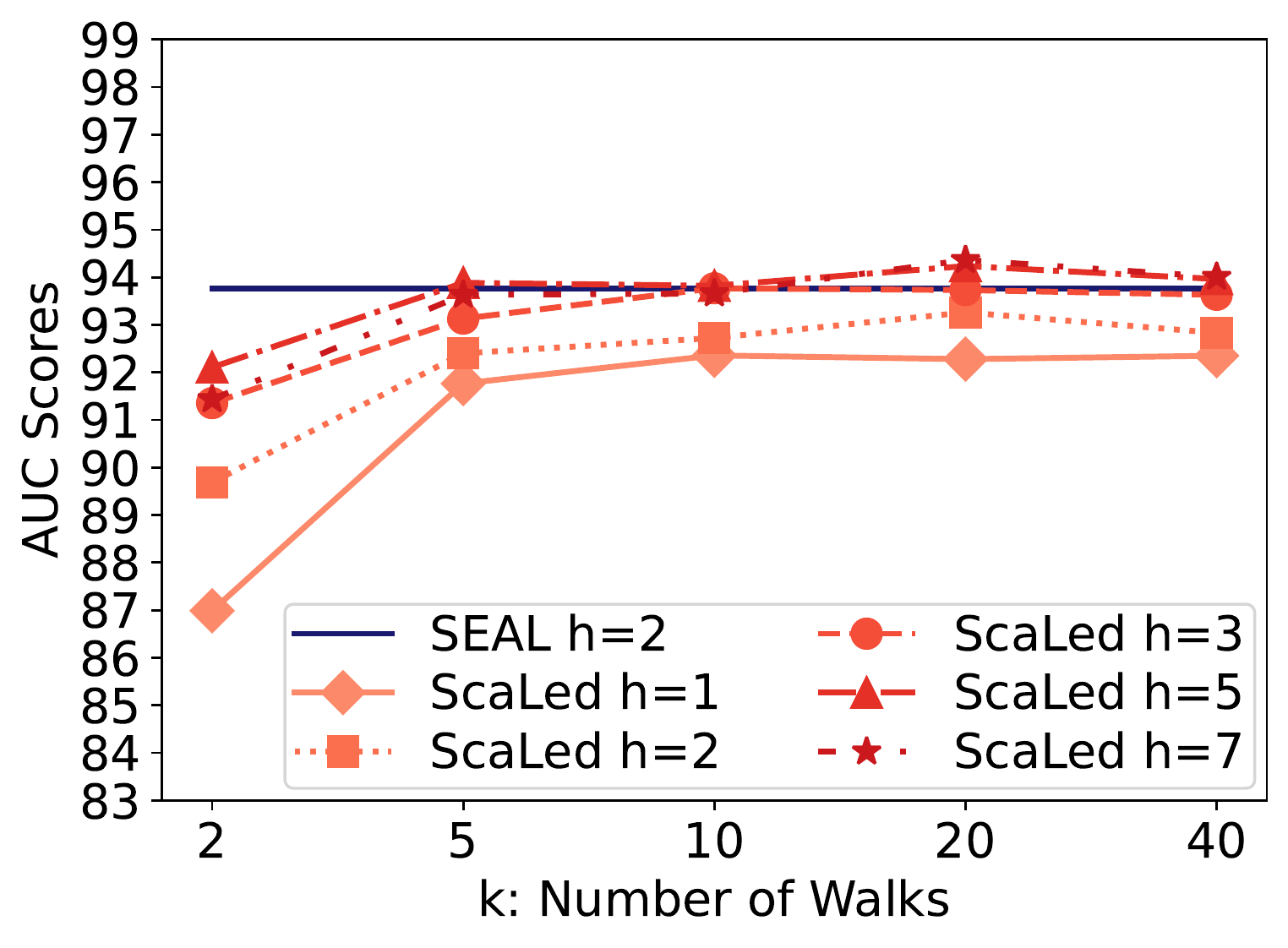}}\hspace{-4pt}
   \subfigure[Runtime for Router.\label{runtimesubfig4}]{\includegraphics[scale=0.29]{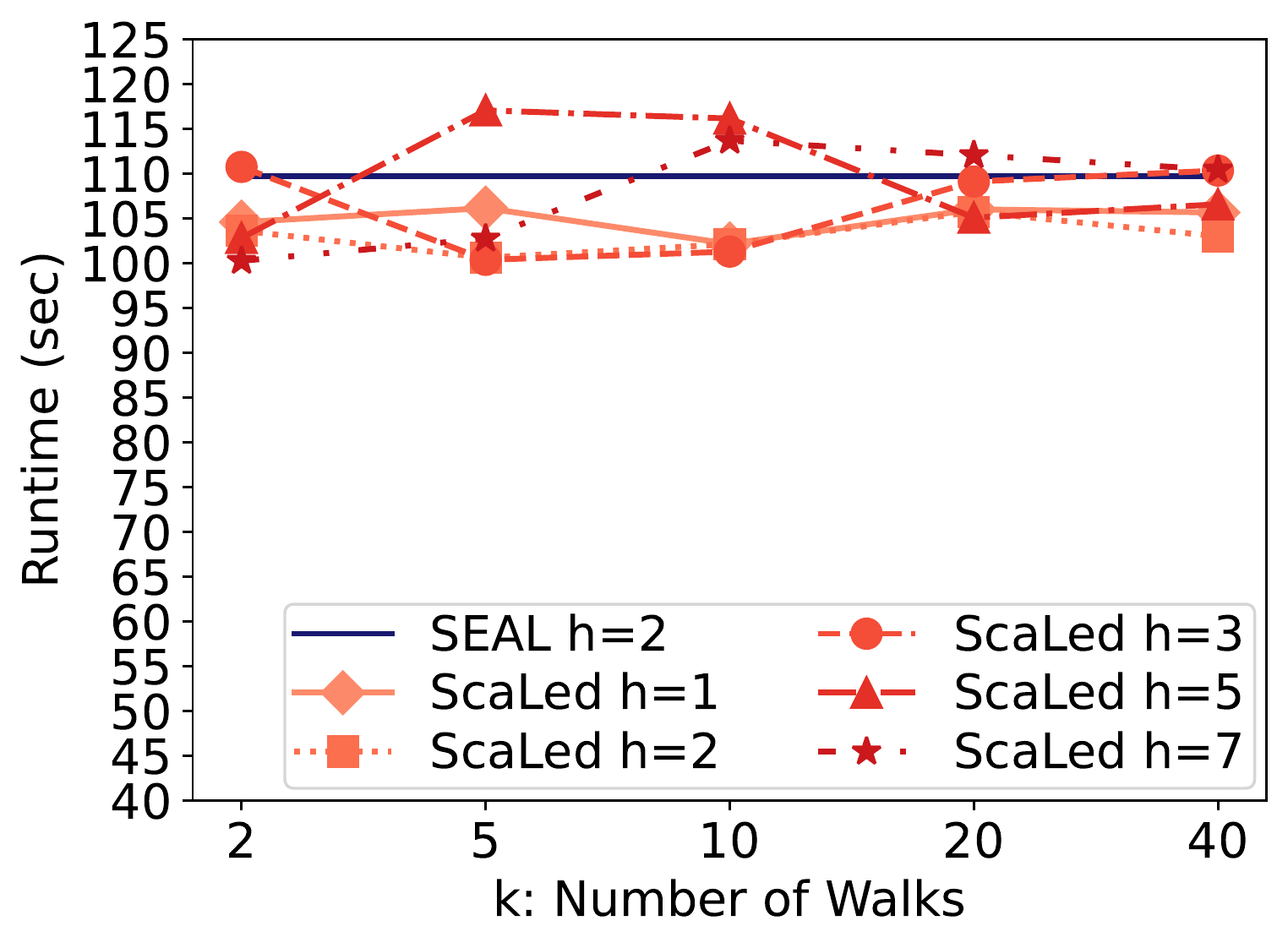}}\hspace{-4pt}
  \subfigure[AUC for Power.\label{subfig5}]{\includegraphics[scale=0.29]{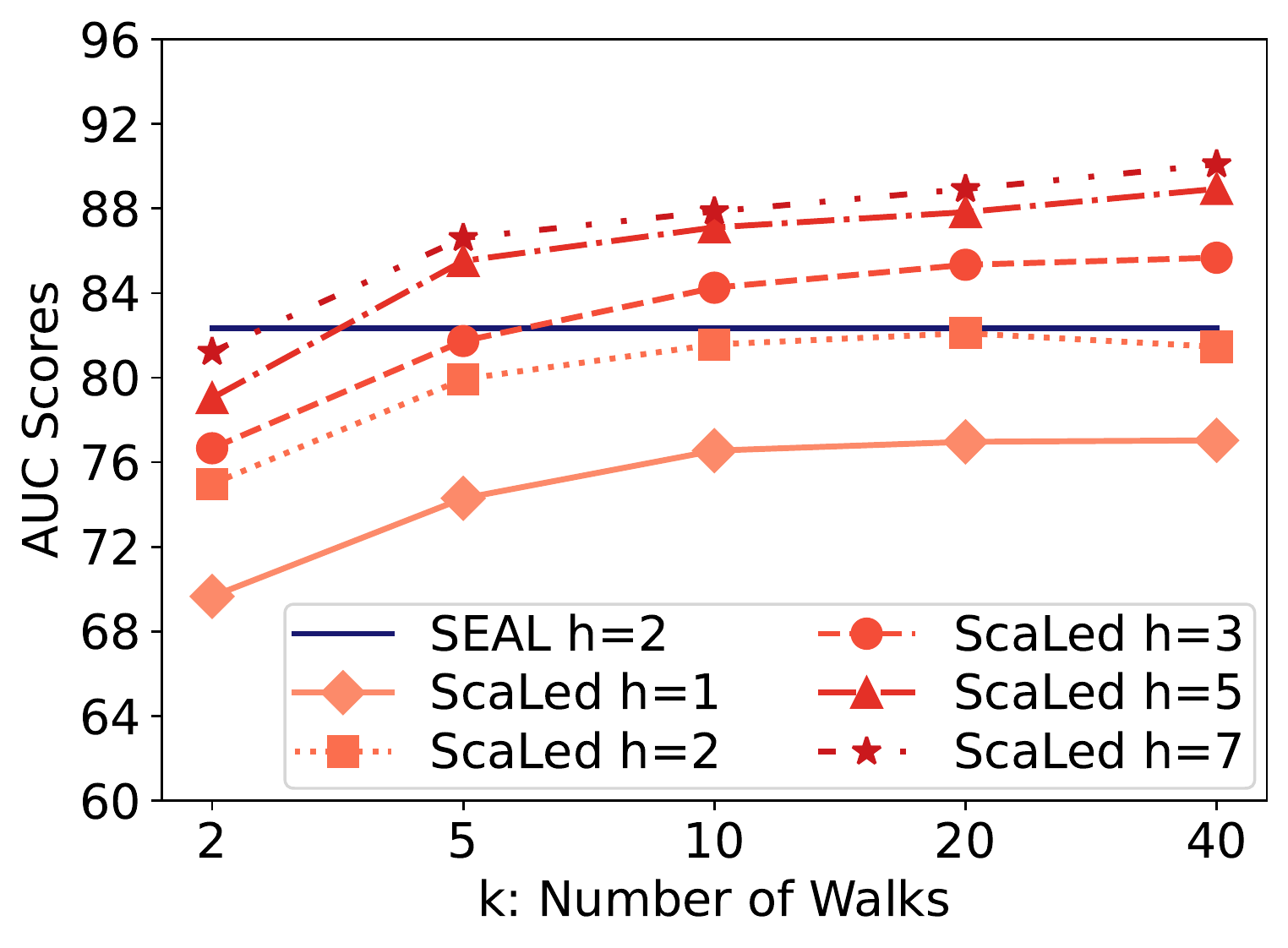}}\hspace{-4pt}
   \subfigure[Runtime for Power.\label{runtimesubfig5}]{\includegraphics[scale=0.29]{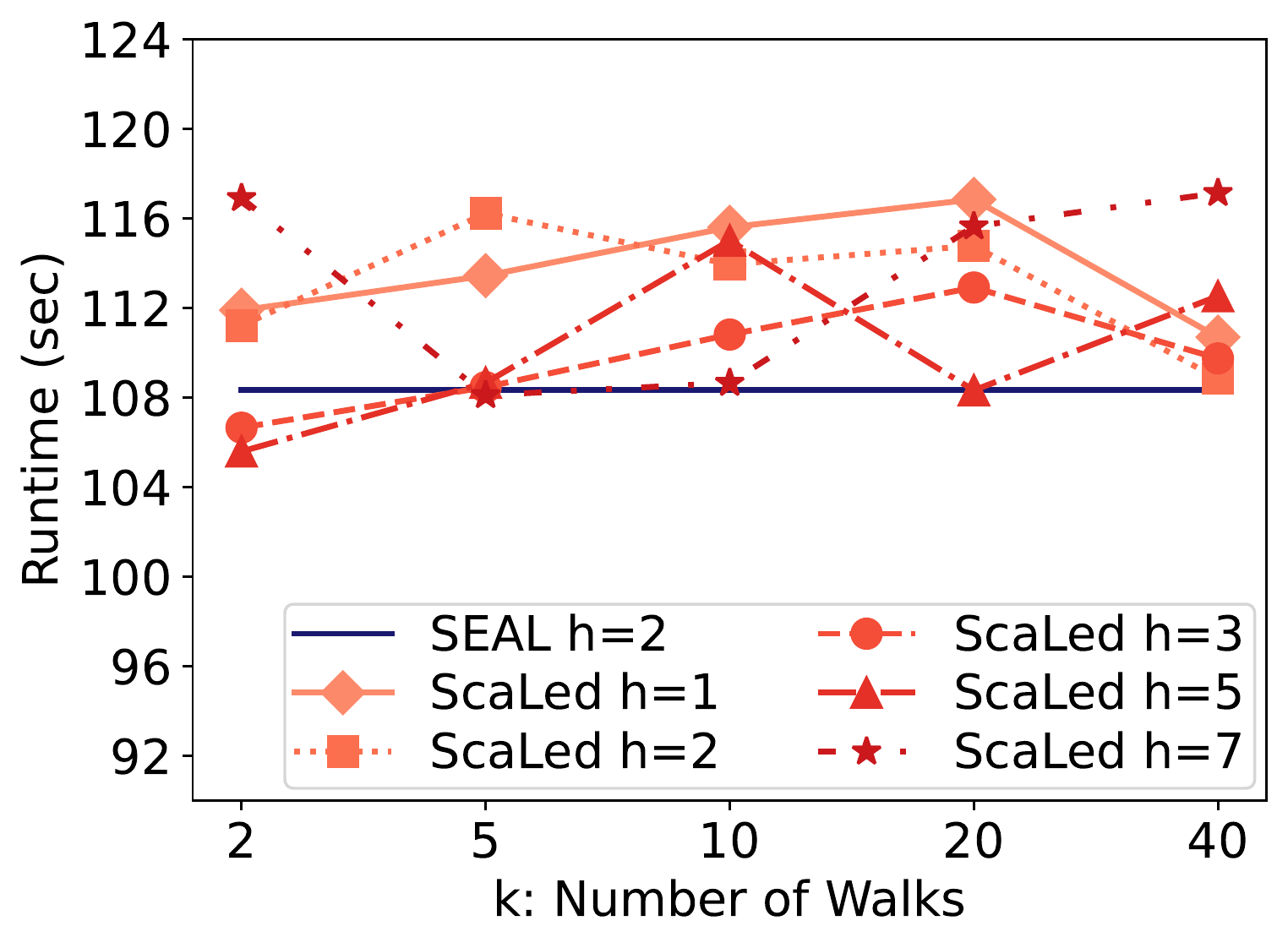}}\hspace{-4pt}
  \subfigure[AUC for Yeast.\label{subfig6}]{\includegraphics[scale=0.29]{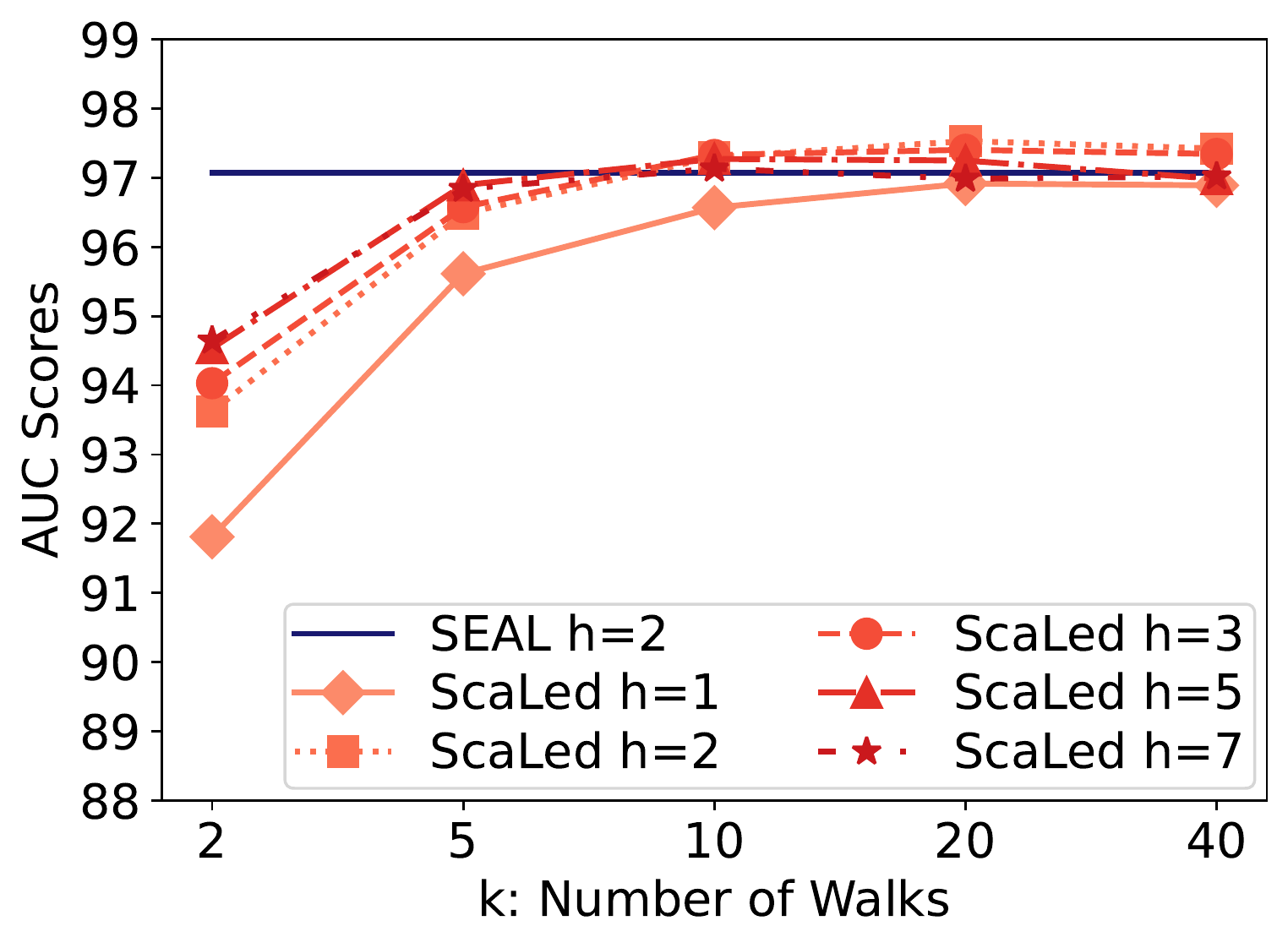}}\hspace{-4pt}
    \subfigure[Runtime for Yeast.\label{runtimesubfig6}]{\includegraphics[scale=0.29]{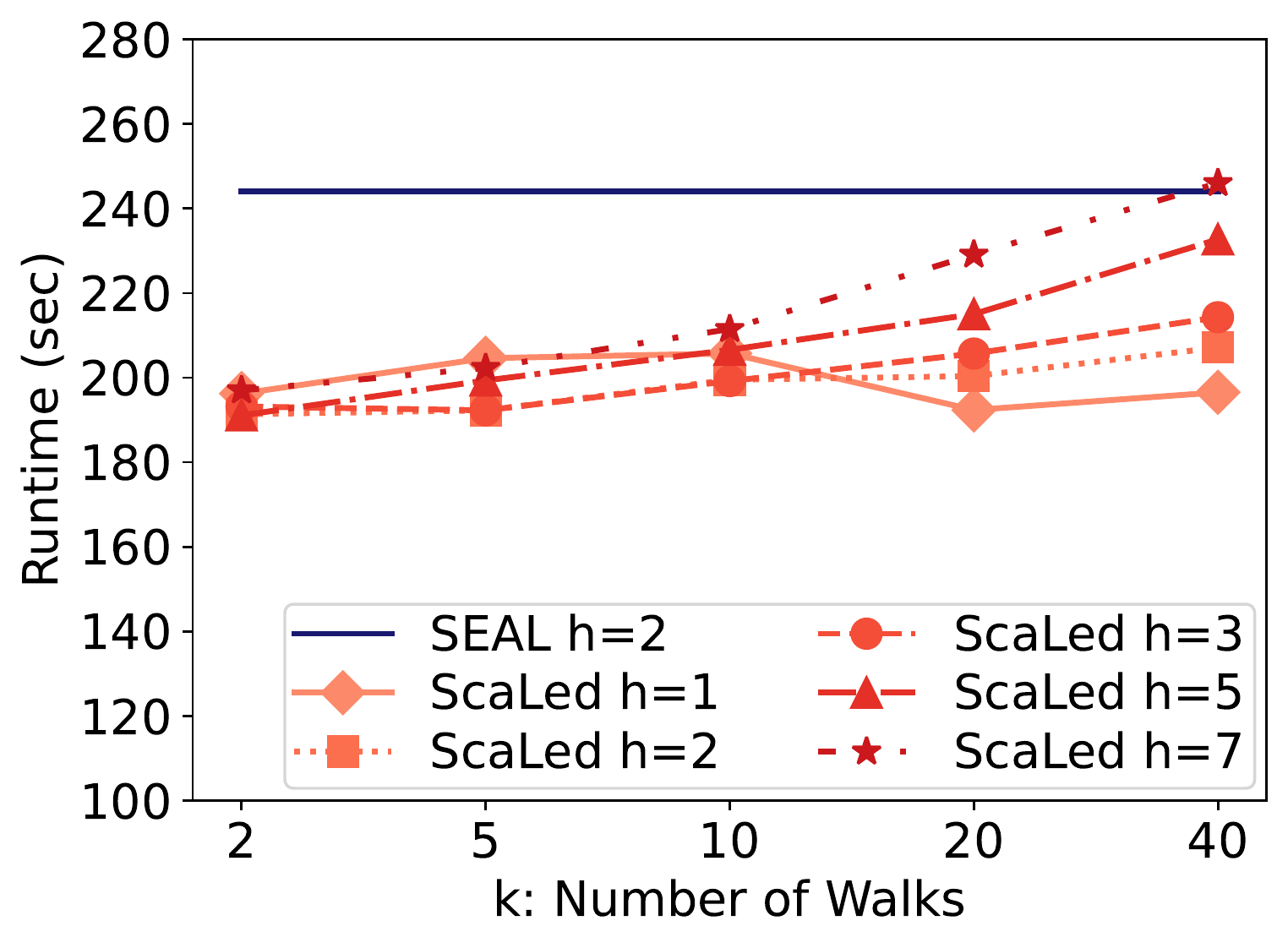}}\hspace{-4pt}
  \subfigure[AUC for Ecoli.\label{subfig7}]{\includegraphics[scale=0.29]{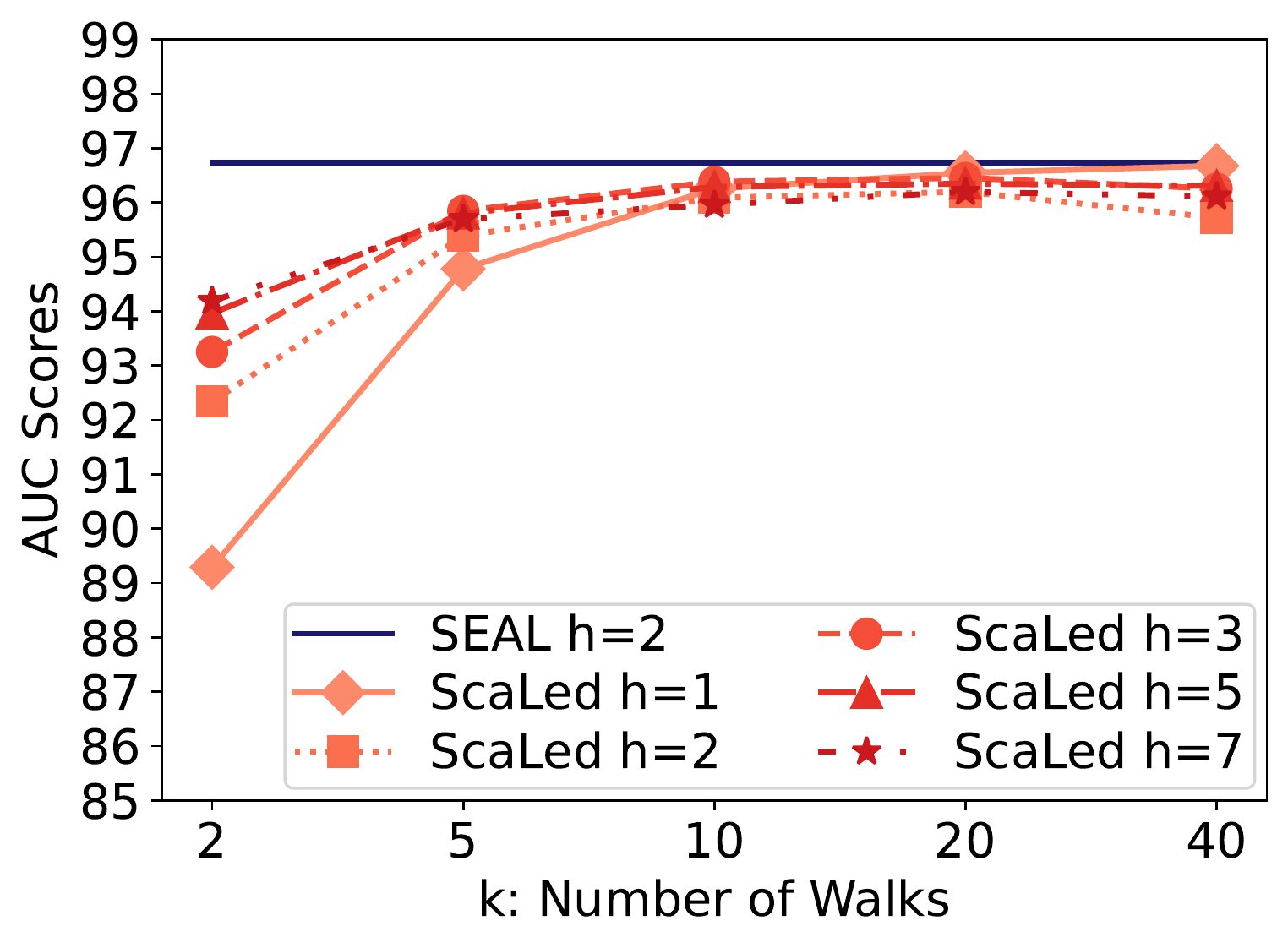}}\hspace{-4pt}
  \subfigure[Runtime for Ecoli.\label{runtimesubfig7}]{\includegraphics[scale=0.29]{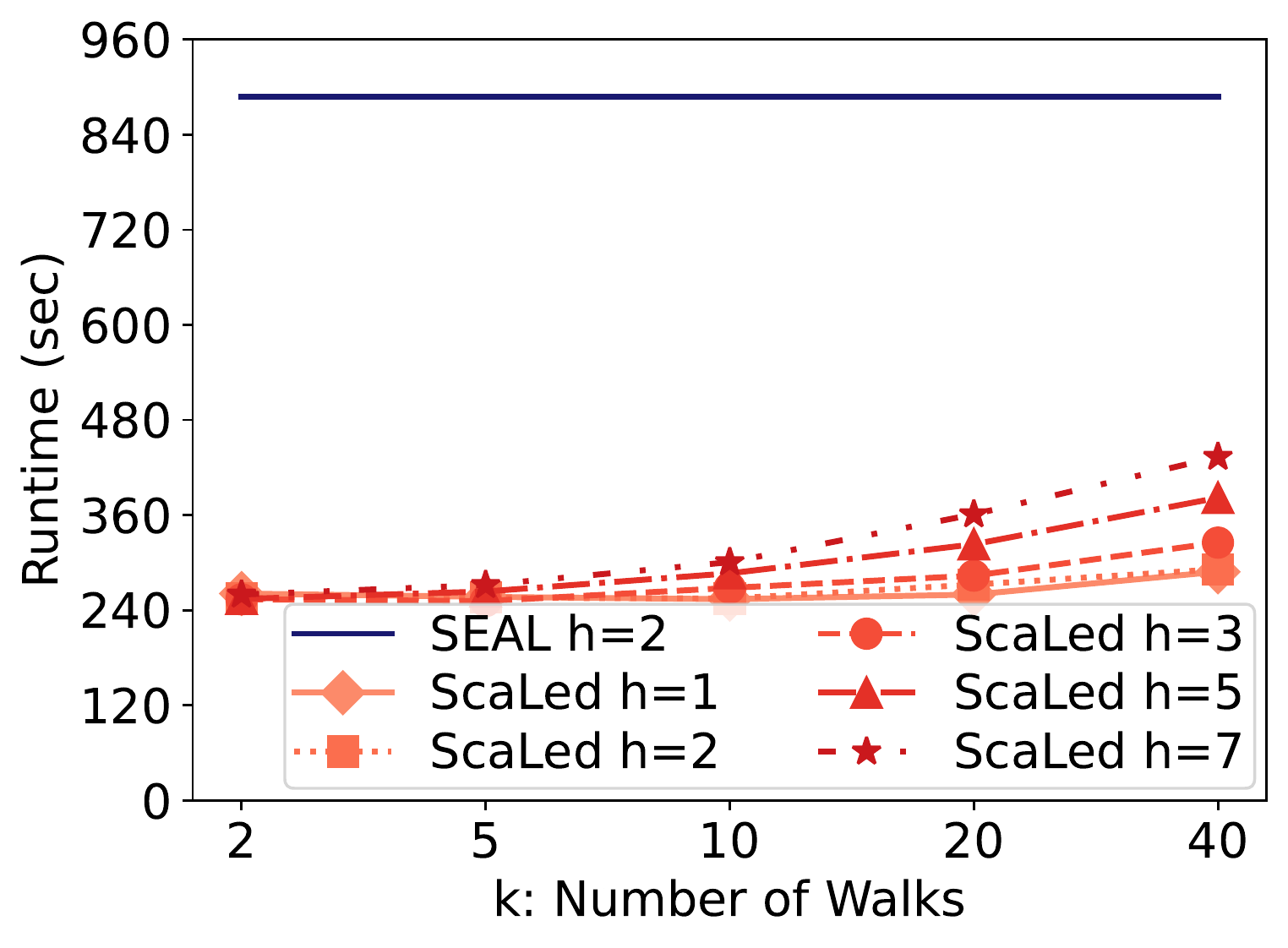}}\hspace{-4pt}
  \subfigure[AUC for PB.\label{subfig8}]{\includegraphics[scale=0.29]{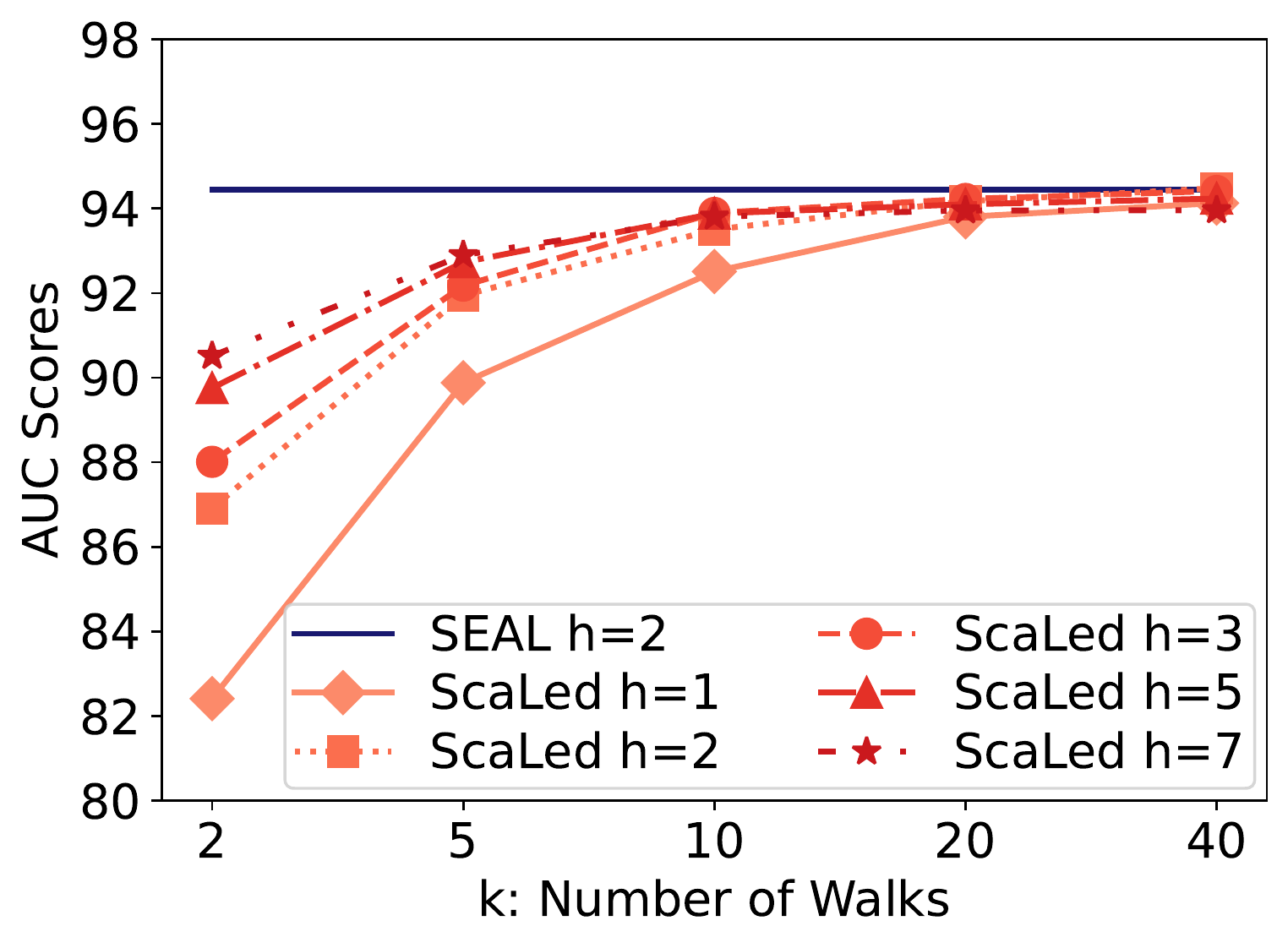}}\hspace{-4pt}
  \subfigure[Runtime for PB.\label{runtimesubfig8}]{\includegraphics[scale=0.29]{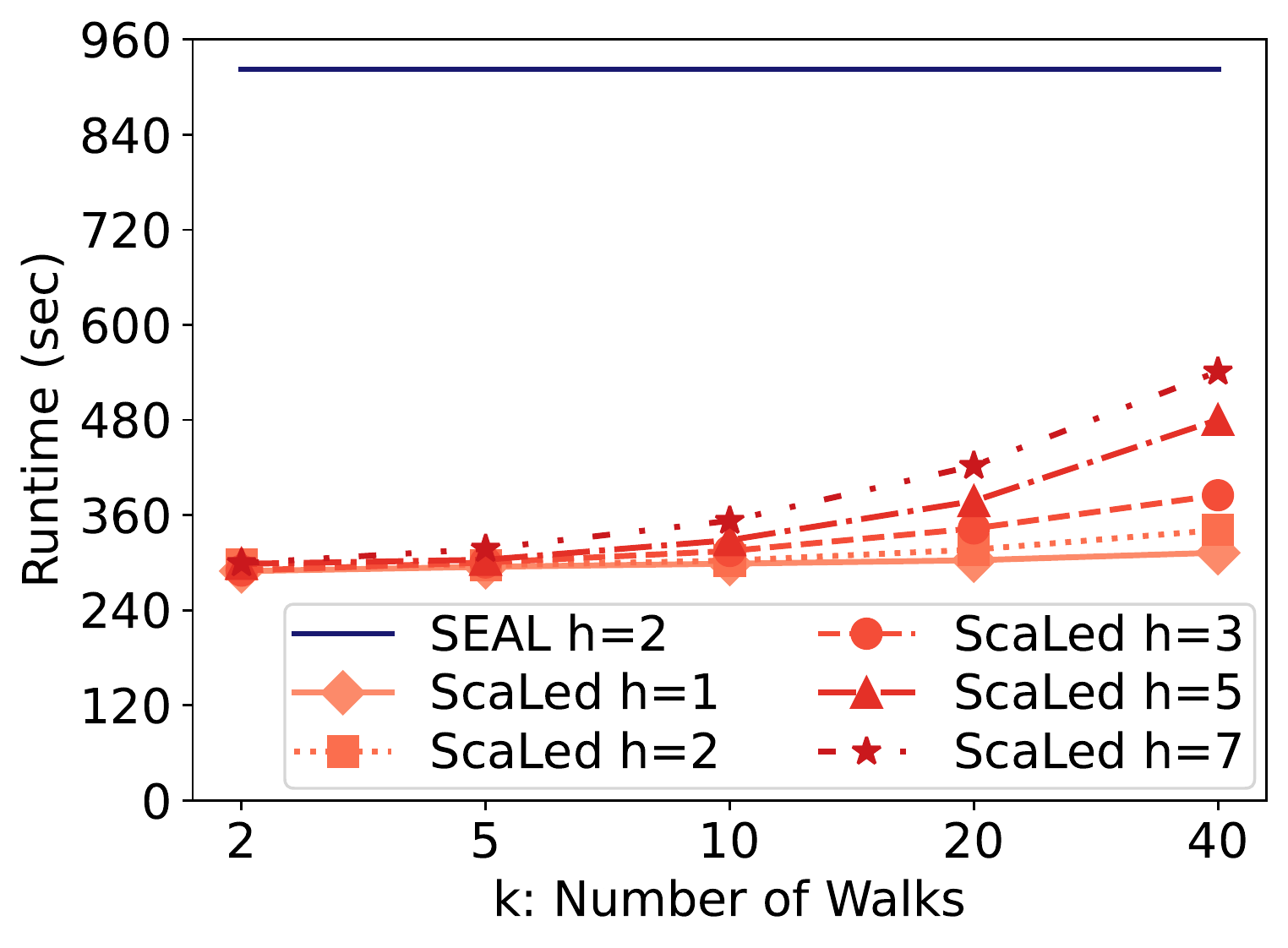}}\hspace{-4pt}
   \subfigure[AUC for Cora.\label{subfig9}]{\includegraphics[scale=0.29]{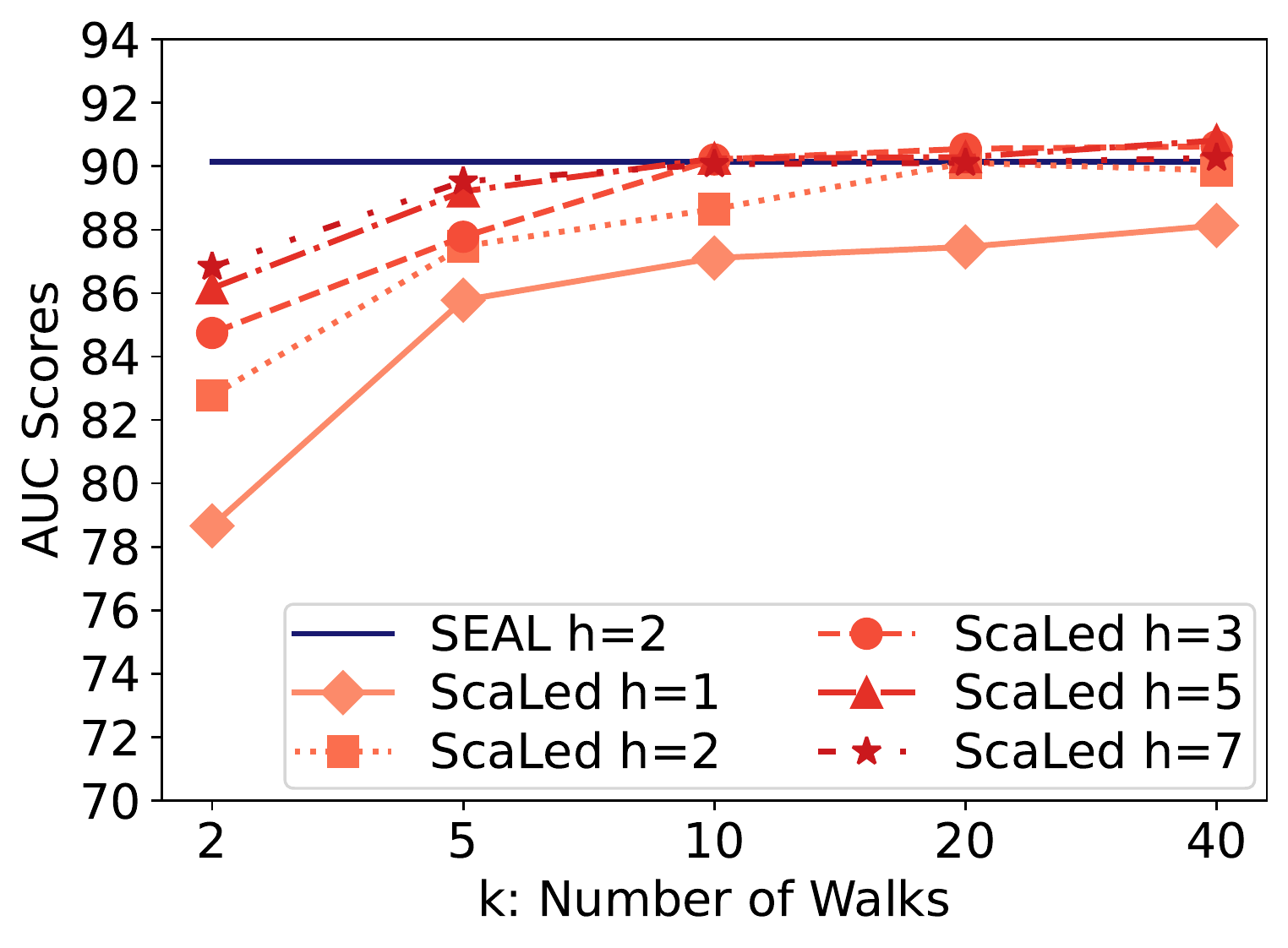}}\hspace{-4pt}
     \subfigure[Runtime for Cora.\label{runtimesubfig9}]{\includegraphics[scale=0.29]{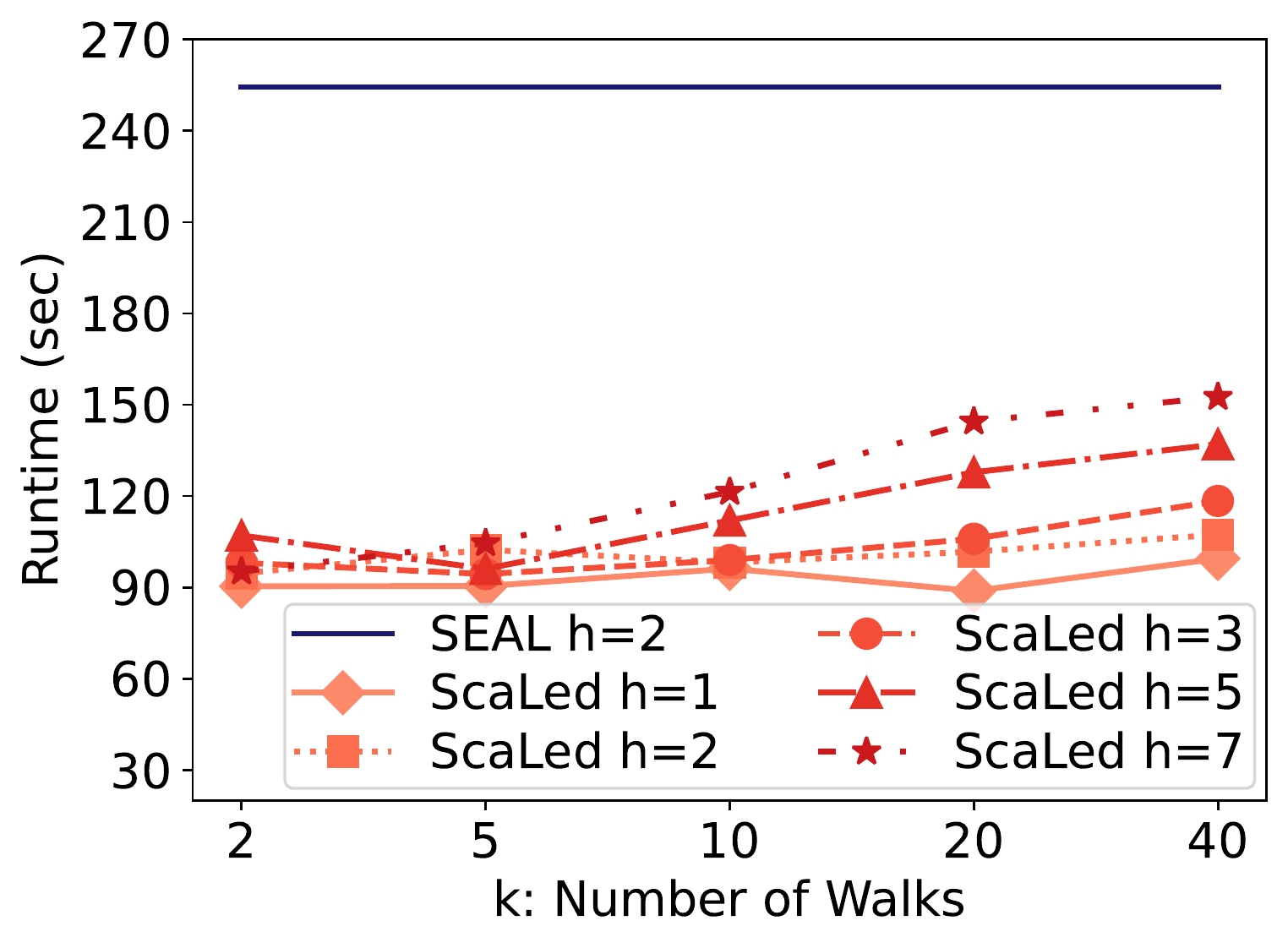}}
  \hspace{-4pt}
  \subfigure[AUC for CiteSeer.\label{subfig10}]{\includegraphics[scale=0.29]{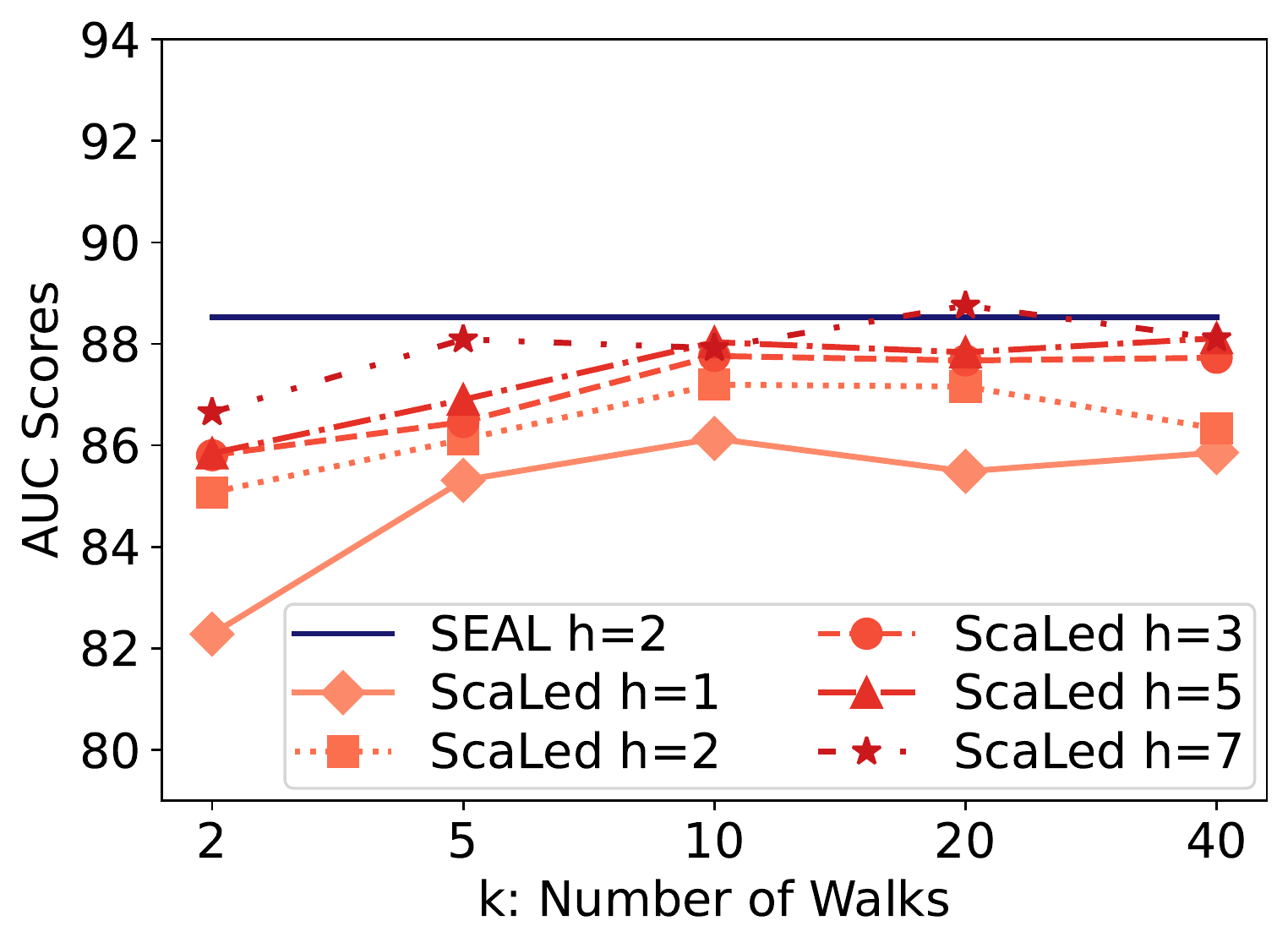}}\hspace{-4pt}
   \subfigure[Runtime for CiteSeer.\label{runtimesubfig10}]{\includegraphics[scale=0.29]{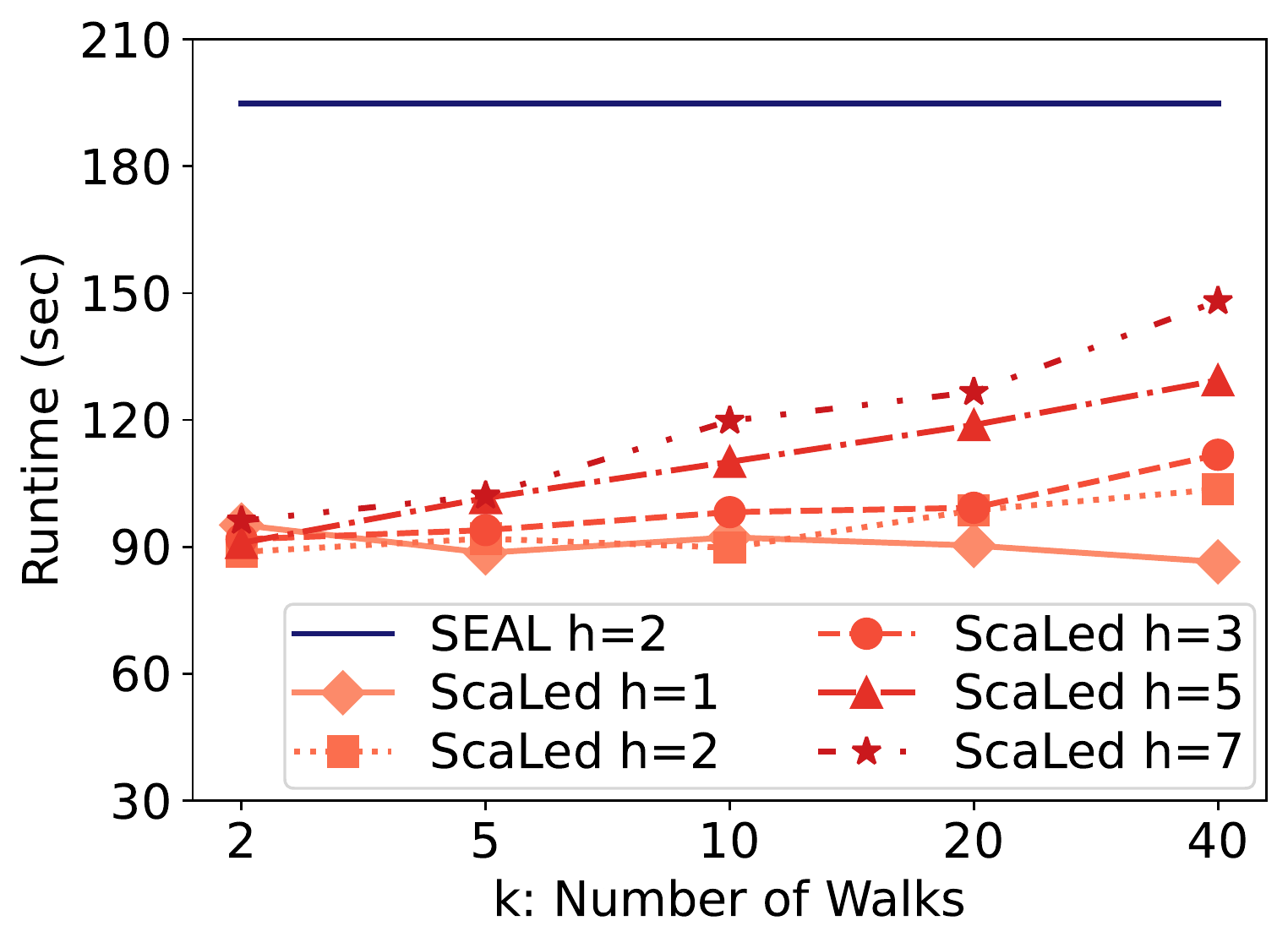}}\hspace{-4pt}
  \caption{Average of AUCs and runtimes over 5 random seeds with various datasets, number of walks $k$, and walk lengh $h$.}
  \label{fig:tuning-results}
  \vspace{-8pt}
\end{figure*}

\vskip 1.5mm \noindent \textbf{Results: Hyperparameter Sensitivity Analyses.} We intend to understand how the walk length $h$ and the number of walks $k$ control the computational overhead in \scaled during training and inference. Thus, we conduct a sensitivity analysis of these two parameters on all of the attributed and non-attributed datasets. We vary $h$ and $k$ while keeping other hyperparameters fixed. Figure \ref{fig:tuning-results} reports the average AUC and runtime for all the datasets over 5 runs.\footnote{For these experiments, we have trained each model for 5 epochs (rather than our default of 50 epochs) in each run. Thus, these results are not comparable with those of Table \ref{table:accuracy_results}.} The AUC results for all datasets are qualitatively similar where the AUC increases with both values of $h$ and $k$. One can also noticeably observe the runtime slightly increases with both walk length $h$ and the number of walks $k$ (with a few exceptions due to statistical noise). But, this slight increment of computational overhead elevates \scaled's accuracy measure and in the case of USAir, NS, Router, Power, Yeast and Cora pushes it towards and beyond that of SEAL. 
Interestingly, for datasets such as Ecoli and PB, when $k$ reaches $20$ and any $h$,  the AUC of \scaled  gets very close to that of SEAL. We also observe that AUC increases much faster with the number of walks $k$  in comparison to the walk length $h$. For $h=2$ and $k=40$, \scaled has outperformed or is comparable to SEAL in terms of AUC, but gained up to $3\times$ speed up, with the exceptions of NS, Router and Power where the runtime is similar to SEAL. The lack of speedup in those datasets might be explained by their low average degrees. Since these graphs are relatively sparse, the random walks discover the (close-to) exact enclosing subgraphs even with some potential redundancy by revisiting discovered nodes. Another interesting observation is that for large $h$ (e.g., $h=7$), \scaled has not been able to outperform SEAL drastically. This observation confirms that a node's local neighborhood has more information and we get diminishing returns by moving farther away from the target nodes. One practical conclusion is that for reaching high accuracy and maximum speed up, one is better off keeping the walk length $h$ low but increasing the number of walks $k$.

\section{Conclusion and Future Works}
Link prediction has emerged as an important task for graph-structured data with applications spanning across multiple domains. Existing state-of-the-art link predicion methods use subgraph representation learning (SGRL), which learns the enriched embedding of the enclosing subgraphs around the pair of nodes. However, SGRL methods are not scalable to large real-world graphs. We proposed \scaled to overcome this scalability shortcomings, which exploits random walks to sample sparser enclosing subgraphs to reduce the computation overhead. The main idea is to preserve the key structural information of subgraphs with less number of nodes and edges, thus yielding smaller computational graphs for GNNs which in turn reduces the runtime and memory consumption. Our extensive experiments demonstrate \scaled can match the accuracy measures of the state-of-the-art link prediction models while  consuming order of magnitudes less resources. For future work, we plan to explore how to adaptively choose the length of the walks and the number of walks depending on the structural positions of two nodes. Another interesting research direction that could be explored is to apply graph augmentation techniques to the sampled subgraphs in \scaled to further enhance its learning capabilities.

\bibliographystyle{ACM-Reference-Format}
\bibliography{arxiv}


\end{document}